\documentclass[10pt,twocolumn,letterpaper]{article}

\usepackage{cvpr}
\usepackage{times}
\usepackage{epsfig}
\usepackage{graphicx}
\usepackage{amsmath}
\usepackage{amssymb}

\usepackage{bm}
\usepackage{booktabs}

\newcommand{\quotes}[1]{``#1''}
\newcommand{\cname}[1]{{\fontfamily{cmtt}\selectfont {#1}}}

\usepackage[pagebackref=true,breaklinks=true,letterpaper=true,colorlinks,bookmarks=false]{hyperref}

\cvprfinalcopy % *** Uncomment this line for the final submission

\def\cvprPaperID{2792} % *** Enter the CVPR Paper ID here

\ifcvprfinal\fi
\begin{document}

%-------------------------------------------------------------------------
% Title
\title{Neural 3D Mesh Renderer}

\author{
    Hiroharu Kato${}^\text{1}$, Yoshitaka Ushiku${}^\text{1}$, and Tatsuya Harada${}^\text{1,2}$\\
    ${}^\text{1}$The University of Tokyo, ${}^\text{2}$RIKEN\\
    {\tt\small \{kato,ushiku,harada\}@mi.t.u-tokyo.ac.jp}
}

\maketitle
%\thispagestyle{empty}

%-------------------------------------------------------------------------
% Abstract
\begin{abstract}
    For modeling the 3D world behind 2D images, which 3D representation is most appropriate? A polygon mesh is a promising candidate for its compactness and geometric properties. However, it is not straightforward to model a polygon mesh from 2D images using neural networks because the conversion from a mesh to an image, or rendering, involves a discrete operation called rasterization, which prevents back-propagation. Therefore, in this work, we propose an approximate gradient for rasterization that enables the integration of rendering into neural networks. Using this renderer, we perform single-image 3D mesh reconstruction with silhouette image supervision and our system outperforms the existing voxel-based approach. Additionally, we perform gradient-based 3D mesh editing operations, such as 2D-to-3D style transfer and 3D DeepDream, with 2D supervision for the first time. These applications demonstrate the potential of the integration of a mesh renderer into neural networks and the effectiveness of our proposed renderer.
\end{abstract}

%-------------------------------------------------------------------------
% Introduction
\section{Introduction}
Understanding the 3D world from 2D images is one of the fundamental problems in computer vision. Humans model the 3D world in their brains using images on their retinas, and live their daily existence using the constructed model. The machines, too, can act more intelligently by explicitly modeling the 3D world behind 2D images.

The process of generating an image from the 3D world is called {\it rendering}. Because this lies on the border between the 3D world and 2D images, it is crucially important in computer vision.

In recent years, convolutional neural networks (CNNs) have achieved considerable success in 2D image understanding~\cite{girshick2014rich,krizhevsky2012imagenet}. Therefore, incorporating rendering into neural networks has a high potential for 3D understanding.

\begin{figure}[t]
    \begin{center}
    \includegraphics[width=1.0\linewidth,bb=0 0 623 481]{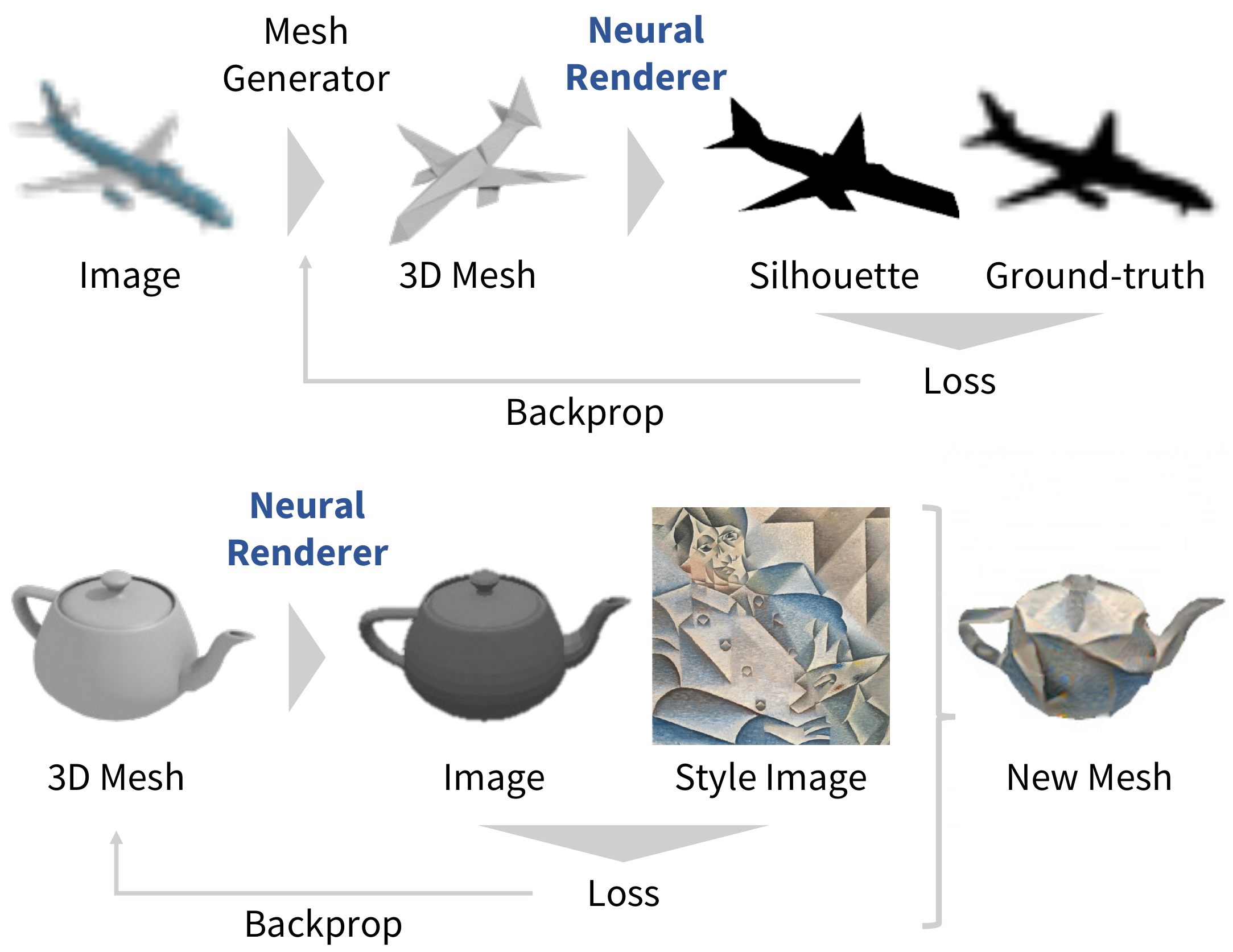}
    \end{center}
       \caption{Pipelines for single-image 3D mesh reconstruction (upper) and 2D-to-3D style transfer (lower).}
    \label{fig:applications}
    \vspace{-2mm}
\end{figure}

What type of 3D representation is most appropriate for modeling the 3D world? Commonly used 3D formats are voxels, point clouds and polygon meshes. Voxels, which are 3D extensions of pixels, are the most widely used format in machine learning because they can be processed by CNNs~\cite{choy20163d,maturana2015voxnet,qi2016volumetric,riegler2016octnet,tatarchenko2017octree,tulsiani2017multi,wu2016learning,wu20153d,yan2016perspective}. However, it is difficult to process high resolution voxels because they are regularly sampled from 3D space and their memory efficiency is poor. The scalability of point clouds, which are sets of 3D points, is relatively high because point clouds are based on irregular sampling. However, textures and lighting are difficult to apply because point clouds do not have surfaces. Polygon meshes, which consist of sets of vertices and surfaces, are promising because they are scalable and have surfaces. Therefore, in this work, we use the polygon mesh as our 3D format.

% new paragraph
One advantage of polygon meshes over other representations in 3D understanding is its compactness. For example, to represent a large triangle, a polygon mesh only requires three vertices and one face, whereas voxels and point clouds require many sampling points over the face. Because polygon meshes represent 3D shapes with a small number of parameters, the model size and dataset size for 3D understanding can be made smaller.

Another advantage is its suitability for geometric transformations. The rotation, translation, and scaling of objects are represented by simple operations on the vertices. This property also facilitates to train 3D understanding models.

% new paragraph
Can we train a system including rendering as a neural network? This is a challenging problem. Rendering consists of projecting the vertices of a mesh onto the screen coordinate system and generating an image through regular grid sampling~\cite{marschner2015fundamentals}. Although the former is a differentiable operation, the latter, referred to as {\it rasterization}, is difficult to integrate because back-propagation is prevented by the discrete operation.

% new paragraph
Therefore, to enable back-propagation with rendering, we propose an approximate gradient for rendering peculiar to neural networks, which facilitates end-to-end training of a system including rendering. Our proposed renderer can flow gradients into texture, lighting, and cameras as well as object shapes. Therefore, it is applicable to a wide range of problems. We name our renderer {\it Neural Renderer}.

% new paragraph
In the generative approach in computer vision and machine learning, problems are solved by modeling and inverting the process of data generation. Images are generated via rendering from the 3D world, and a polygon mesh is an efficient, rich and intuitive 3D representation. Therefore, \quotes{backward pass} of mesh renderers is extremely important.

In this work, we propose the two applications illustrated in Figure~\ref{fig:applications}. The first is single-image 3D mesh reconstruction with silhouette image supervision. Although 3D reconstruction is one of the main problems in computer vision, there are few studies to reconstruct meshes from single images despite the potential capacity of this approach. The other application is gradient-based 3D mesh editing with 2D supervision. This includes a 3D version of style transfer~\cite{gatys2016image} and DeepDream~\cite{mordvintsev2015inceptionism}. This task cannot be realized without a differentiable mesh renderer because voxels or point clouds have no smooth surfaces.

The major contributions can be summarized as follows.
\begin{itemize}
    \setlength\itemsep{0em}
    \item We propose an approximate gradient for rendering of a mesh, which enables the integration of rendering into neural networks.
    \item We perform 3D mesh reconstruction from single images without 3D supervision and demonstrate our system's advantages over the voxel-based approach.
    \item We perform gradient-based 3D mesh editing operations, such as 2D-to-3D style transfer and 3D DeepDream, with 2D supervision for the first time.
    \item We will release the code for Neural Renderer.
\end{itemize}

%-------------------------------------------------------------------------
\section{Related work}
In this section, we briefly describe how 3D representations have been integrated into neural networks. We also summarize works related to our two applications.

%-----------------------------------------------------------------
\subsection{3D representations in neural networks}
3D representations are categorized into rasterized and geometric forms. Rasterized forms include voxels and multi-view RGB(D) images. Geometric forms include point clouds, polygon meshes, and sets of primitives.

Rasterized forms are widely used because they can be processed by CNNs. Voxels, which are 3D extensions of pixels, are used for classification~\cite{maturana2015voxnet,qi2016volumetric,riegler2016octnet,wu2016learning,wu20153d}, 3D reconstruction and generation~\cite{choy20163d,tatarchenko2017octree,tulsiani2017multi,wu2016learning,yan2016perspective}. Because the memory efficiency of voxels is poor, some recent works have incorporated more efficient representations~\cite{riegler2016octnet,tatarchenko2017octree,wang2017cnn}. Multi-view RGB(D) images, which represent a 3D scene through a set of images, are used for recognition~\cite{qi2016volumetric,su2015multi} and view synthesis~\cite{tatarchenko2016multi}.

Geometric forms require some modifications to be integrated into neural networks. For example, systems that handle point clouds must be invariant to the order of points. Point clouds have been used for both recognition~\cite{klokov2017escape,qi2017pointnet,qi2017pointnetplusplus} and reconstruction~\cite{fan2016point}. Primitive-based representations, which represent 3D objects using a set of primitives, such as cuboids, have also been investigated~\cite{li2017grass,zou20173d}.

A Polygon mesh represents a 3D object as a set of vertices and surfaces. Because it is memory efficient, suitable for geometric transformations, and has surfaces, it is the de facto standard form in computer graphics (CG) and computer-aided design (CAD). However, because the data structure of a polygon mesh is a complicated graph, it is difficult to integrate into neural networks. Although recognition and segmentation have been investigated~\cite{kalogerakis20163d,yi2016syncspeccnn}, generative tasks are much more difficult. Rezende \etal~\cite{rezende2016unsupervised} incorporated the OpenGL renderer into a neural network for 3D mesh reconstruction. Gradients of the black-box renderer were estimated using REINFORCE~\cite{williams1992simple}. In contrast, the gradients in our renderer are geometry-grounded and presumably more accurate. OpenDR~\cite{loper2014opendr} is a differentiable renderer. Unlike this general-purpose renderer, our proposed gradients are designed for neural networks.

%-----------------------------------------------------------------
\subsection{Single-image 3D reconstruction}
The estimation of 3D structures from images is a traditional problem in computer vision. Following the recent progress in machine learning algorithms, 3D reconstruction from a single image has become an active research topic.

Most methods learn a 2D-to-3D mapping function using ground truth 3D models. While some works reconstruct 3D structures via depth prediction~\cite{eigen2014depth,saxena20083}, others directly predict 3D shapes~\cite{choy20163d,fan2016point,tatarchenko2017octree,tulsiani2017multi,wu2016learning}.

Single-image 3D reconstruction can be realized without 3D supervision. Perspective transformer nets (PTN)~\cite{yan2016perspective} learn 3D structures using silhouette images from multiple viewpoints. Our 3D reconstruction method is also based on silhouette images. However, we use polygon meshes whereas they used voxels.

%-----------------------------------------------------------------
\subsection{Image editing via gradient descent}
Using a differentiable feature extractor and loss function, an image that minimizes the loss can be generated via back-propagation and gradient descent. DeepDream~\cite{mordvintsev2015inceptionism} is an early example of such a system. An initial image is repeatedly updated so that the magnitude of its image feature becomes larger. Through this procedure, objects such as dogs and cars gradually appear in the image.

Image style transfer~\cite{gatys2016image} is likely the most familiar and practical example. Given a {\it content image} and {\it style image}, an image with the specified content and style is generated.

% new paragraph
Our renderer provides gradients of an image with respect to the vertices and textures of a mesh. Therefore, DeepDream and style transfer of a mesh can be realized by using loss functions on 2D images.

%-------------------------------------------------------------------------
\section{Approximate gradient for rendering}

In this section, we describe Neural Renderer, which is a 3D mesh renderer with gradient flow.

\begin{figure}[t]
    \begin{center}
    \includegraphics[width=1.0\linewidth]{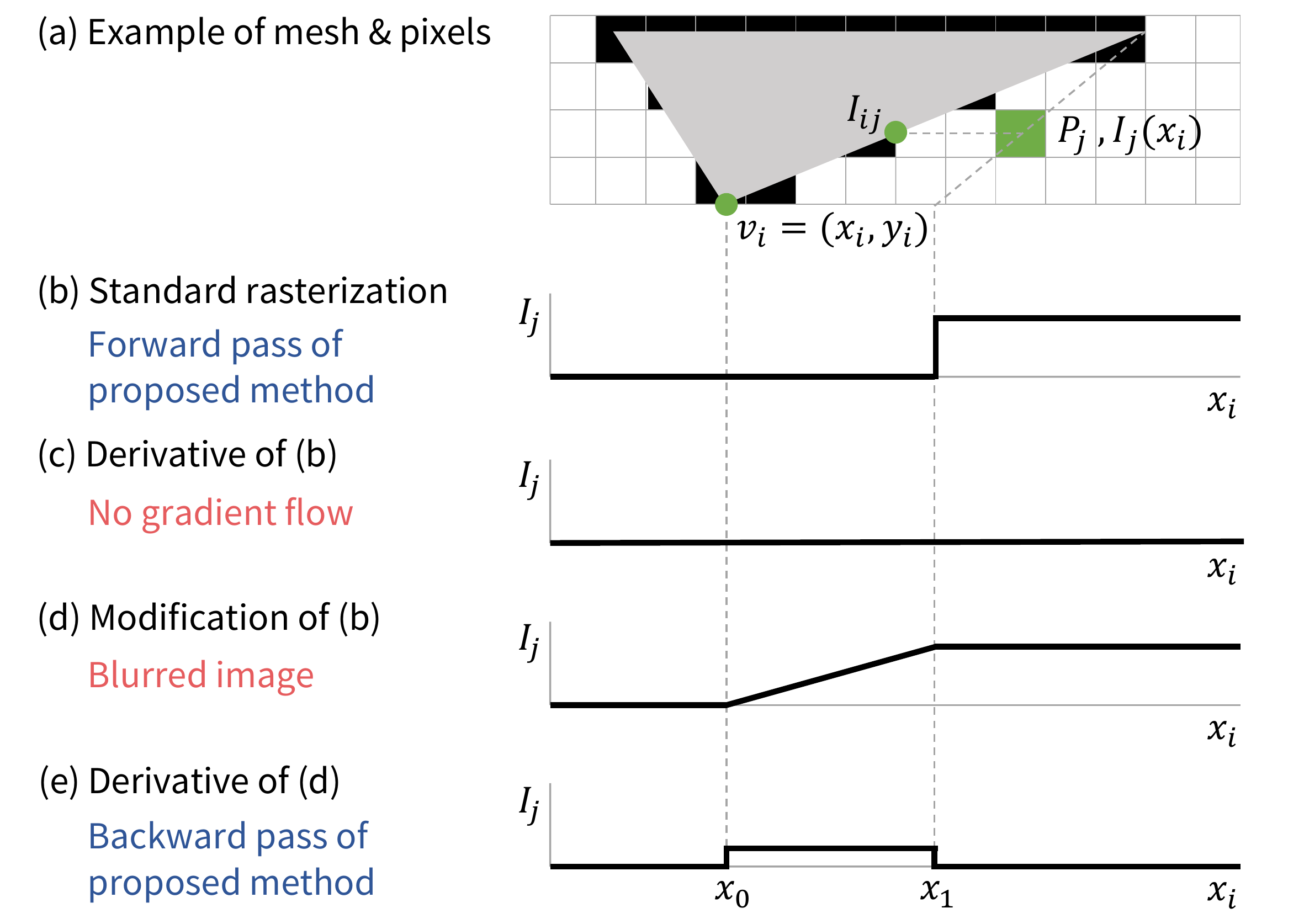}
    \end{center}
       \caption{Illustration of our method. $\bm{v}_i = \{x_i, y_i\}$ is one vertex of the face. $I_j$ is the color of pixel $P_j$. The current position of $x_i$ is $x_0$. $x_1$ is the location of $x_i$ where an edge of the face collides with the center of $P_j$ when $x_i$ moves to the right. $I_j$ becomes $I_{ij}$ when $x_i=x_1$.}
    \label{fig:method}
    \vspace{-2mm}
\end{figure}

%-----------------------------------------------------------------
\subsection{Rendering pipeline and its derivative}

A 3D mesh consists of a set of vertices $\{ \bm{v}^{o}_{1}, \bm{v}^{o}_{2}, .., \bm{v}^{o}_{N_v} \}$ and faces $\{ \bm{f}_1, \bm{f}_2, .., \bm{f}_{N_f} \}$, where the object has $N_v$ vertices and $N_f$ faces. $\bm{v}^{o}_i \in \mathbb{R}^3$ represents the position of the $i$-th vertex in the 3D object space and $\bm{f}_j \in \mathbb{N}^3$ represents the indices of the three vertices corresponding to the $j$-th triangle face. To render this object, vertices $\{ \bm{v}_{i}^{o} \}$ in the object space are transformed into vertices $\{ \bm{v}_{i}^{s} \}$, $\bm{v}_{i}^{s} \in \mathbb{R}^2$ in the screen space. This transformation is represented by a combination of differentiable transformations~\cite{marschner2015fundamentals}.

An image is generated from $\{ \bm{v}_i^s \}$ and $\{ \bm{f}_j \}$ via sampling. This process is called rasterization. Figure~\ref{fig:method}~(a) illustrates rasterization in the case of single triangle. If the center of a pixel $P_j$ is inside of the face, the color $I_j$ of the pixel $P_j$ becomes the color of the overlapping face $I_{ij}$. Because this is a discrete operation, assuming that $I_{ij}$ is independent of $\bm{v}_i$, $\frac{\partial I_j}{\partial \bm{v}_i}$ is zero almost everywhere, as shown in Figure~\ref{fig:method}~(b--c). This means that the error signal back-propagated from a loss function to pixel $P_j$ does not flow into the vertex $\bm{v}_{i}$.

\begin{figure}[t]
    \begin{center}
    \includegraphics[width=1.0\linewidth]{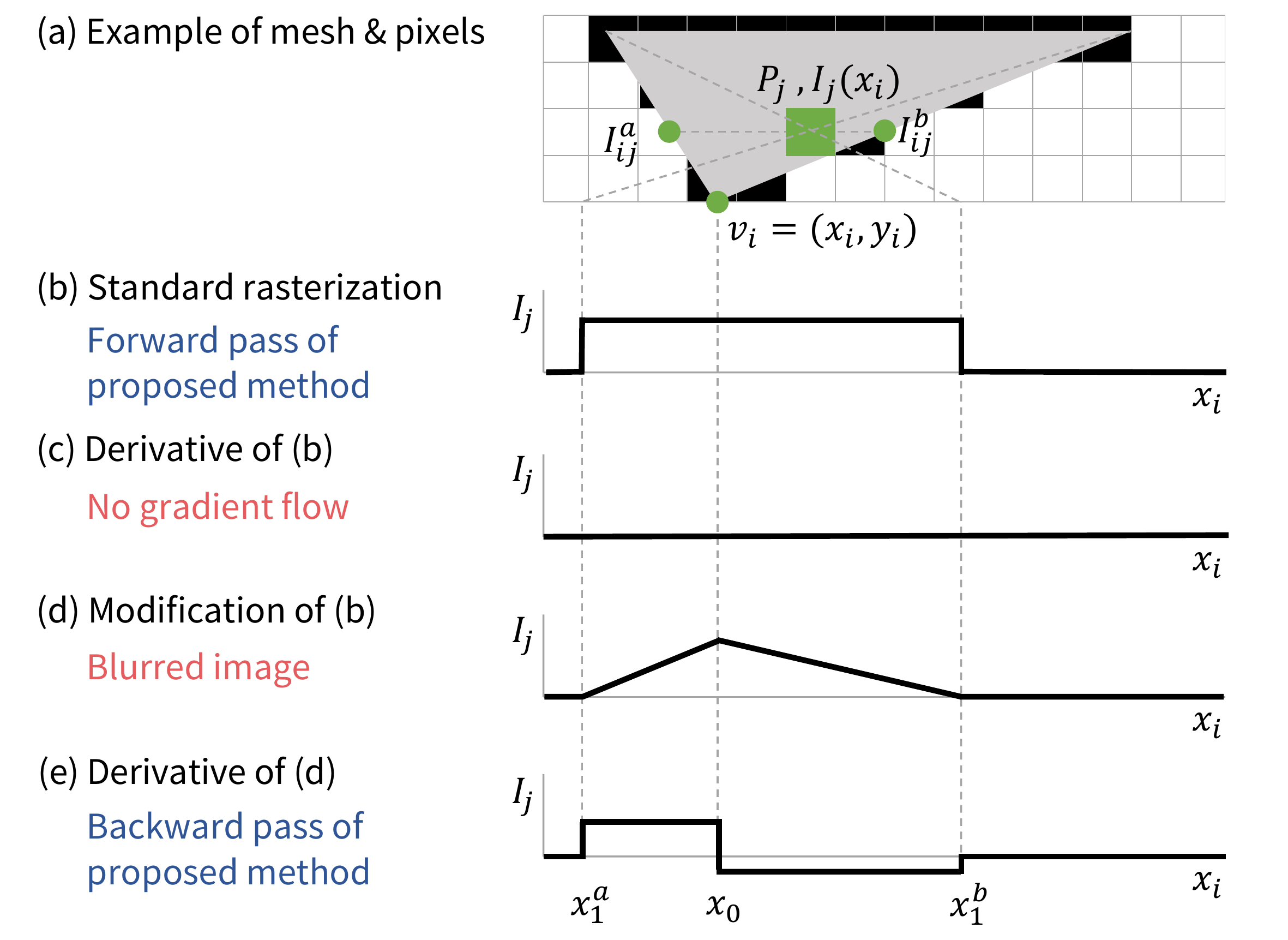}
    \end{center}
       \caption{Illustration of our method in the case where $P_j$ is inside the face. $I_j$ changes when $x_i$ moves to the right or left.}
    \label{fig:method2}
    \vspace{-2mm}
\end{figure}

%-----------------------------------------------------------------
\subsection{Rasterization of a single face}

For ease of explanation, we describe our method using the x-coordinate $x_i$ of a single vertex $\bm{v}_i=\bm{v}_i^s$ in the screen space and a single gray-scale pixel $P_j$. We consider the color of $P_j$ to be a function $I_j(x_i)$ on $x_i$ and freeze all variables other than $x_i$.

First, we assume that $P_j$ is outside the face, as shown in Figure~\ref{fig:method}~(a). The color of $P_j$ is $I(x_0)$ when $x_i$ is at the current position $x_0$. If $x_i$ moves to the right and reaches the point $x_1$, where an edge of the face collides with the center of $P_j$, $I_j(x_i)$ suddenly turns to the color of hitting point $I_{ij}$. Let $\delta^x_i$ be the distance traveled by $x_i$, let $\delta^x_i = x_1 - x_0$, and let $\delta^I_j$ represent the change in the color $\delta^I_j = I(x_1) - I(x_0)$. The partial derivative $\frac{\partial I_j(x_i)}{\partial x_i}$ is zero almost everywhere, as illustrated in Figure~\ref{fig:method}~(b--c).

Because the gradient is zero, the information that $I_j(x_0)$ can be changed by $\delta^I_j $ if $x_i$ moves by $\delta^x_i$ to the right is not transmitted to $x_i$. This is because $I_j(x_i)$ suddenly changes. Therefore, we replace the sudden change with a gradual change between $x_0$ and $x_1$ using linear interpolation. Then, $\frac{\partial I_j}{\partial x_i}$ becomes $\frac{\delta^I_j}{\delta^x_i}$ between $x_0$ and $x_1$, as shown in Figure~\ref{fig:method}~(d--e).

The derivative of $I_j(x_i)$ is different on the right and left sides of $x_0$. How should one define a derivative at $x_i=x_0$? We propose switching the values using the error signal $\delta_j^P$ back-propagated to $P_j$. The sign of $\delta_j^P$ indicates whether $P_j$ should be brighter or darker. To minimize the loss, if $\delta_j^P > 0$, then $P_j$ must be darker. On the other hand, the sign of $\delta^I_j$ indicates whether $P_j$ can be brighter or darker. If $\delta^I_j > 0$, $P_j$ becomes brighter by pulling in $x_i$, but $P_j$ cannot become darker by moving $x_i$. Therefore, a gradient should not flow if $\delta_j^P > 0$ and $\delta^I_j > 0$. From this viewpoint, we define $\frac{\partial I_j(x_i)}{\partial x_i} \rvert_{x_i=x_0}$ as follows.
\begin{eqnarray}
    &\frac{\partial I_j(x_i)}{\partial x_i} \Bigr|_{x_i=x_0} =
    \begin{cases}
        \frac{\delta^I_j}{\delta^x_i}; &\delta_j^P \delta^I_j < 0.\\
        0; &\delta_j^P \delta^I_j \geq 0.
    \end{cases}
\end{eqnarray}

Sometimes, the face does not overlap $P_j$ regardless of where $x_i$ moves. This means that $x_1$ does not exist. In this case, we define $\frac{\partial I_j(x_i)}{\partial x_i}\rvert_{x_i=x_0} = 0$.

We use Figure~\ref{fig:method}~(b) for the forward pass because if we use Figure~\ref{fig:method}~(d), the color of a face leaks outside of the face. Therefore, our rasterizer produces the same images as the standard rasterizer, but it has non-zero gradients.

The derivative with respect to $y_i$ can be obtained by swapping the x-axis and y-axis in the above discussion.

Next, we consider a case where $P_j$ is inside the face, as shown in Figure~\ref{fig:method2} (a). In this case, $I(x_i)$ changes when $x_i$ moves to the right or left. Standard rasterization, its derivative, an interpolated function, and its derivative are shown in Figure~\ref{fig:method2} (b--e). We first compute the derivatives on the left and right sides of $x_0$ and let their sum be the gradient at $x_0$. Specifically, using the notation in Figure~\ref{fig:method2}, $\delta^{I^a}_j=I(x_1^a)-I(x_0)$, $\delta^{I^b}_j=I(x_1^b)-I(x_0)$, $\delta_x^a=x_1^a-x_0$ and $\delta_x^b=x_1^b-x_0$, we define the loss as follows.
\begin{eqnarray}
    &\frac{\partial I_j(x_i)}{\partial x_i} \Bigr|_{x_i=x_0} = \frac{\partial I_j(x_i)}{\partial x_i} \Bigr|_{x_i=x_0}^a + \frac{\partial I_j(x_i)}{\partial x_i} \Bigr|_{x_i=x_0}^b. \\
    &\frac{\partial I_j(x_i)}{\partial x_i} \Bigr|_{x_i=x_0}^a =
    \begin{cases}
        \frac{\delta^{I^a}_j}{\delta_x^a}; &\delta_j^P \delta^{I^a}_j < 0.\\
        0; &\delta_j^P \delta^{I^a}_j \geq 0.\\
    \end{cases}\\
    &\frac{\partial I_j(x_i)}{\partial x_i} \Bigr|_{x_i=x_0}^b =
    \begin{cases}
        \frac{\delta^{I^b}_j}{\delta_x^b}; &\delta_j^P \delta^{I^b}_j < 0.\\
        0; &\delta_j^P \delta^{I^a}_j \geq 0.\\
    \end{cases}
\end{eqnarray}

%-----------------------------------------------------------------
\subsection{Rasterization of multiple faces}
If there are multiple faces, our rasterizer draws only the frontmost face at each pixel, which is the same as the standard method~\cite{marschner2015fundamentals}. During the backward pass, we first check whether or not the cross points $I_{ij}$, $I_{ij}^a$, and $I_{ij}^b$ are drawn, and do not flow gradients if they are occluded by surfaces not including $\bm{v}_i$.

%-----------------------------------------------------------------
\subsection{Texture}
Textures can be mapped onto faces. In our implementation, each face has its own texture image of size $s_t \times s_t \times s_t$. We determine the coordinates in the texture space corresponding to a position $\bm{p}$ on a triangle $\{ \bm{v}_1, \bm{v}_2, \bm{v}_3 \}$ using the centroid coordinate system. In other words, if $\bm{p}$ is expressed as $\bm{p} = w_1 \bm{v}_1 + w_2 \bm{v}_2 + w_3 \bm{v}_3$, let $(w_1, w_2, w_3)$ be the corresponding coordinates in the texture space. Bilinear interpolation is used for sampling from a texture image.

%-----------------------------------------------------------------
\subsection{Lighting}
Lighting can be applied directly to a mesh, unlike voxels and point clouds. In this work, we use a simple ambient light and directional light without shading. Let $l^a$ and $l^d$ be the intensities of the ambient light and directional light, respectively, $\bm{n}^d$ be a unit vector indicating the direction of the directional light, and $\bm{n}_j$ be the normal vector of a surface. We then define the modified color of a pixel $I^l_j$ on the surface as $I^l_j = \left( l^a + \left( \bm{n}^d \cdot \bm{n}_j \right) l^d \right) I_j $.

In this formulation, gradients also flow into the intensities $l^a$ and $l^d$, as well as the direction $\bm{n}^d$ of the directional light. Therefore, light sources can also be included as an optimization target.

%-------------------------------------------------------------------------
\section{Applications of Neural Renderer}
We apply our proposed renderer to (a) single-image 3D reconstruction with silhouette image supervision and (b) gradient-based 3D mesh editing, including a 3D version of style transfer~\cite{gatys2016image} and DeepDream~\cite{mordvintsev2015inceptionism}. An image of a mesh $m$ rendered from a viewpoint $\phi_i$ is denoted $R(m,\phi_i)$.

%-----------------------------------------------------------------
\subsection{Single image 3D reconstruction}
\label{sec:method_reconstruction}
Yan \etal~\cite{yan2016perspective} demonstrated that single-image 3D reconstruction can be realized without 3D training data. In their setting, a 3D generation function $G(x)$ on an image $x$ was trained such that silhouettes of a predicted 3D shape $\{ \hat{s}_i = R(G(x), \phi_i) \}$ match the ground truth silhouettes $\{ s_i \}$, assuming that the viewpoints $\{ \phi_i \}$ are known. This pipeline is illustrated in Figure~\ref{fig:applications}. While Yan \etal~\cite{yan2016perspective} generated voxels, we generate a mesh.

Although voxels can be generated by extending existing image generators~\cite{goodfellow2014generative,radford2015unsupervised} to the 3D space, mesh generation is not so straightforward. In this work, instead of generating a mesh from scratch, we deform a predefined mesh to generate a new mesh. Specifically, we use an isotropic sphere with $642$ vertices and move each vertex $\bm{v}_i$ as $\bm{v}_i + \bm{b}_i + \bm{c}$ using a local bias vector $\bm{b}_i$ and global bias vector $\bm{c}$. Additionally, we restrict the movable range of each vertex within the same quadrant on the original sphere. The faces $\{ \bm{f}_i \}$ are unchanged. Therefore, the intermediate outputs of $G(x)$ are $\bm{b} \in \mathbb{R}^{642 \times 3}$ and $\bm{c} \in \mathbb{R}^{1 \times 3}$. The mesh we use is specified by $642 \times 3$ parameters, which is far less than the typical voxel representation with a size of $32^3$. This low-dimensionality is presumably beneficial for shape estimation.

The generation function $G(x)$ is trained using silhouette loss $\mathcal{L}_{\textrm{sl}}$ and smoothness loss $\mathcal{L}_{\textrm{sm}}$. Silhouette loss represents how much the reconstructed silhouettes $\{ \hat{s}_i \}$ differ from the correct silhouettes $\{ s_i \}$. Smoothness loss represents how smooth the surfaces of a mesh are and acts as a regularizer. The objective function is a weighted sum of these two loss functions $\mathcal{L} = \lambda_{\textrm{sl}} \mathcal{L}_{\textrm{sl}} + \lambda_{\textrm{sm}} \mathcal{L}_{\textrm{sm}}$.

Let $\{ s_i \}$ and $\{ \hat{s}_i \}$ be binary masks, $\theta_i$ be the angle between two faces including the $i$-th edge in $G(x)$, $\mathcal{E}$ be the set of all edges in $G(x)$, and $\odot$ be an element-wise product. We define the loss functions as:
\begin{eqnarray}
    \mathcal{L}_{\textrm{sl}} (x | \phi_i, s_i) &=& -\frac{ \left| \hat{s}_i \odot s_i \right|_1 }{\left| \hat{s}_i + s_i - \hat{s}_i \odot s_i \right|_1 }. \\
    \mathcal{L}_{\textrm{sm}} (x) &=& \sum_{\theta_i \in \mathcal{E}} (\cos \theta_i + 1)^2.
\end{eqnarray}
$\mathcal{L}_{\textrm{sl}}$ corresponds to a negative intersection over union (IoU) between the true and reconstructed silhouettes. $\mathcal{L}_{\textrm{sm}}$ ensures that intersection angles of all faces are close to $180$ degrees.

We assume that the object region in an image is segmented via preprocessing in common with the exiting works~\cite{fan2016point,tulsiani2017multi,yan2016perspective}. We input the mask of the object region into the generator as an additional channel of an RGB image.

%-----------------------------------------------------------------
\subsection{Gradient-based 3D mesh editing}
Gradient-based image editing techniques~\cite{gatys2016image,mordvintsev2015inceptionism} generate an image by minimizing a loss function $\mathcal{L}(x)$ on a 2D image $x$ via gradient descent. In this work, instead of generating an image, we optimize a 3D mesh $m$ consisting of vertices $\{ \bm{v}_i \}$, faces $\{ \bm{f}_i \}$, and textures $\{ \bm{t}_i \}$ based on its rendered image $R(m|\phi_i)$.

%-----------------------------------------------------------------
\subsubsection{2D-to-3D style transfer}

In this section, we propose a method to transfer the style of an image $x^s$ onto a mesh $m^c$.

For 2D images, style transfer is achieved by minimizing {\it content loss} and {\it style loss} simultaneously~\cite{gatys2016image}. Specifically, content loss is defined using a feature extractor $f_c(x)$ and content image $x^c$ as $\mathcal{L}_{c}(x|x^c) = \left| f_c(x) - f_c(x^c) \right|^2_2$. Style loss is defined using another feature extractor $f_s(x)$ and style image $x^s$ as $\mathcal{L}_{s}(x|x^s) = \left| M(f_s(x)) - M(f_s(x^s)) \right|^2_{F}$. $M(x)$ transforms a vector into a Gram matrix.

In 2D-to-3D style transfer, content is specified as a 3D mesh $m^c$. To make the shape of the generated mesh similar to that of $m^c$, assuming that the vertices-to-faces relationships $\{ \bm{f}_i \}$ are the same for both meshes, we redefine content loss as $\mathcal{L}_{c} (m | m^c) = \sum_{ \{ \bm{v}_i, \bm{v}^c_i \} \in (m, m^c) } \left| \bm{v}_i - \bm{v}^c_i \right|^2_2$. We use the same style loss as that in the 2D application. Specifically, $\mathcal{L}_{s}(m | x^s, \phi) = \left| M(f_s(R(m,\phi))) - M(f_s(x_s)) \right|^2_{F}$. We also use a regularizer for noise reduction. Let $\mathcal{P}$ denote the a set of colors of all pairs of adjacent pixels in an image $R(m,\phi)$. We define this loss as $\mathcal{L}_{t}(m|\phi) = \sum_{\{ \bm{p}_a, \bm{p}_b \} \in \mathcal{P}} \left| \bm{p}_a - \bm{p}_b \right|_2^2$.

The objective function is $\mathcal{L}=\lambda_c \mathcal{L}_c  + \lambda_s \mathcal{L}_s + \lambda_t \mathcal{L}_t$. We set an initial solution of $m$ as $m^c$ and minimize $\mathcal{L}$ with respect to $\{ \bm{v}_i \}$ and $\{ \bm{t}_i \}$.

%-----------------------------------------------------------------
\subsubsection{3D DeepDream}

Let $f(x)$ be a function that outputs a feature map of an image $x$. For 2D images, a DeepDream of image $x_0$ is achieved by minimizing $-|f(x)|^2_{F}$ via gradient descent starting from $x=x_0$. Optimization is halted after a few iterations. Following a similar process, we minimize $-|f(R(m,\phi))|^2_{F}$ with respect to $\{ \bm{v}_i \}$ and $\{ \bm{t}_i \}$.

%-------------------------------------------------------------------------
\section{Experiments}
In this section, we evaluate the effectiveness of our renderer through the two applications.

%-----------------------------------------------------------------
\subsection{Single image 3D reconstruction}

\begin{figure*}[t]
    \begin{center}
        \includegraphics[height=0.138\textwidth,width=0.138\textwidth,trim=4 4 4 4,clip]{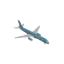}
        \includegraphics[height=0.138\textwidth,width=0.138\textwidth,trim=16 16 16 16,clip]{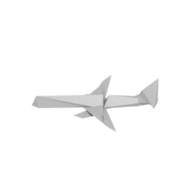}
        \includegraphics[height=0.138\textwidth,width=0.138\textwidth,trim=16 16 16 16,clip]{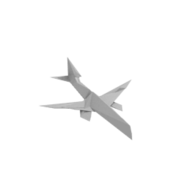}
        \includegraphics[height=0.138\textwidth,width=0.138\textwidth,trim=16 16 16 16,clip]{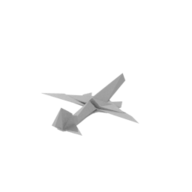}
        \includegraphics[height=0.138\textwidth,width=0.138\textwidth,trim=16 16 16 16,clip]{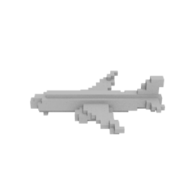}
        \includegraphics[height=0.138\textwidth,width=0.138\textwidth,trim=16 16 16 16,clip]{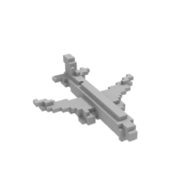}
        \includegraphics[height=0.138\textwidth,width=0.138\textwidth,trim=16 16 16 16,clip]{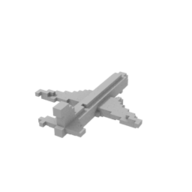} \\
        \includegraphics[height=0.138\textwidth,width=0.138\textwidth,trim=4 4 4 4,clip]{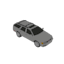}
        \includegraphics[height=0.138\textwidth,width=0.138\textwidth,trim=16 16 16 16,clip]{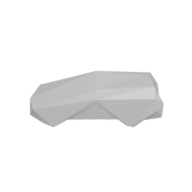}
        \includegraphics[height=0.138\textwidth,width=0.138\textwidth,trim=16 16 16 16,clip]{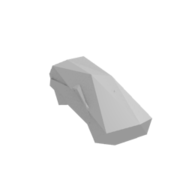}
        \includegraphics[height=0.138\textwidth,width=0.138\textwidth,trim=16 16 16 16,clip]{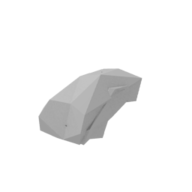}
        \includegraphics[height=0.138\textwidth,width=0.138\textwidth,trim=16 16 16 16,clip]{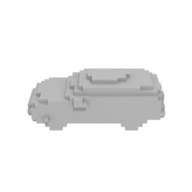}
        \includegraphics[height=0.138\textwidth,width=0.138\textwidth,trim=16 16 16 16,clip]{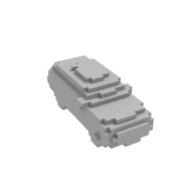}
        \includegraphics[height=0.138\textwidth,width=0.138\textwidth,trim=16 16 16 16,clip]{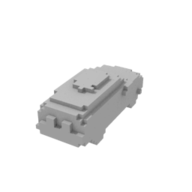} \\
        \includegraphics[height=0.138\textwidth,width=0.138\textwidth,trim=4 4 4 4,clip]{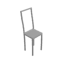}
        \includegraphics[height=0.138\textwidth,width=0.138\textwidth,trim=16 16 16 16,clip]{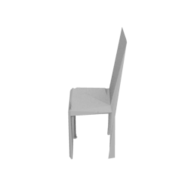}
        \includegraphics[height=0.138\textwidth,width=0.138\textwidth,trim=16 16 16 16,clip]{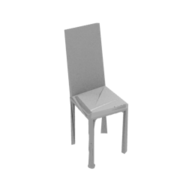}
        \includegraphics[height=0.138\textwidth,width=0.138\textwidth,trim=16 16 16 16,clip]{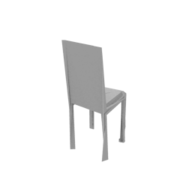}
        \includegraphics[height=0.138\textwidth,width=0.138\textwidth,trim=16 16 16 16,clip]{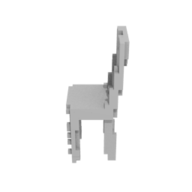}
        \includegraphics[height=0.138\textwidth,width=0.138\textwidth,trim=16 16 16 16,clip]{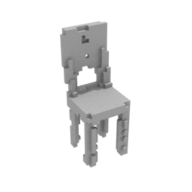}
        \includegraphics[height=0.138\textwidth,width=0.138\textwidth,trim=16 16 16 16,clip]{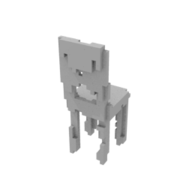} \\
    \end{center}
    \caption{3D mesh reconstruction from a single image. Results are rendered from three viewpoints. First column: input images. Second through fourth columns: mesh reconstruction (proposed method). Fifth through seventh columns: voxel reconstruction~\cite{yan2016perspective}. }
    \label{fig:reconstruction}
    \vspace{-2mm}
\end{figure*}

\begin{table*}[t]
    \begin{center}
        \begin{tabular}{lcccccccc}
            \toprule
            & \cname{airplane} & \cname{bench} & \cname{dresser} & \cname{car} & \cname{chair} & \cname{display} & \cname{lamp} \\
            \hline
            Retrieval~\cite{yan2016perspective} & $0.5564$ & $0.4875$ & $0.5713$ & $0.6519$ & $0.3512$ & $0.3958$ & $0.2905$ \\
            Voxel-based~\cite{yan2016perspective} & $0.5556$ & $0.4924$ & $0.6823$ & $\mathbf{0.7123}$ & $0.4494$ & $0.5395$ & $\mathbf{0.4223}$ \\
            Mesh-based (ours) & $\mathbf{0.6172}$ & $\mathbf{0.4998}$ & $\mathbf{0.7143}$ & $0.7095$ & $\mathbf{0.4990}$ & $\mathbf{0.5831}$ & $0.4126$ \\
            \midrule
            & \cname{loudspeaker} & \cname{rifle} & \cname{sofa} & \cname{table} & \cname{telephone} & \cname{vessel} && \cname{mean} \\
            \hline
            Retrieval~\cite{yan2016perspective} & $0.4600$ & $0.5133$ & $0.5314$ & $0.3097$ & $0.6696$ & $0.4078$ && $0.4766$ \\
            Voxel-based~\cite{yan2016perspective} & $0.5868$ & $0.5987$ & $0.6221$ & $\mathbf{0.4938}$ & $0.7504$ & $0.5507$ && $0.5736$ \\
            Mesh-based (ours) & $\mathbf{0.6536}$ & $\mathbf{0.6322}$ & $\mathbf{0.6735}$ & $0.4829$ & $\mathbf{0.7777}$ & $\mathbf{0.5645}$ && $\mathbf{0.6016}$ \\
            \bottomrule
        \end{tabular}
    \end{center}
    \caption{Reconstruction accuracy measured by voxel IoU. Higher is better. Our mesh-based approach outperforms the voxel-based approach~\cite{yan2016perspective} in $10$ out of $13$ categories.}
    \label{table_reconstruction}
    \vspace{-2mm}
\end{table*}

\begin{figure}[t]
    \begin{center}
        \includegraphics[height=0.138\textwidth,width=0.138\textwidth,trim=4 4 4 4,clip]{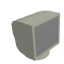}
        \includegraphics[height=0.138\textwidth,width=0.138\textwidth,trim=16 16 16 16,clip]{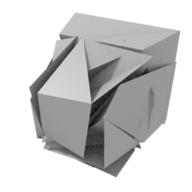}
        \includegraphics[height=0.138\textwidth,width=0.138\textwidth,trim=16 16 16 16,clip]{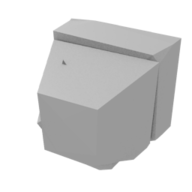} \
    \end{center}
    \caption{Generation of the back side of a CRT monitor with/without smoothness regularizer. Left: input image. Center: prediction without regularizer. Right: prediction with regularizer. }
    \label{fig:reconstruction2}
    \vspace{-2mm}
\end{figure}

%---------------------------------------------------------
\subsubsection{Experimental settings}

% \paragraph{Dataset.}
To compare our mesh-based method with the voxel-based approach by Yan \etal~\cite{yan2016perspective}, we used nearly the same dataset as they did\footnote{The dataset we used was not exactly the same as that used in \cite{yan2016perspective}. The rendering parameters for the input images were slightly different. Additionally, while our silhouette images were rendered by Blender from the meshes in the ShapeNetCore dataset, theirs were rendered by their PTNs using voxelized data.}. We used 3D objects from $13$ categories in the ShapeNetCore~\cite{chang2015shapenet} dataset. Images were rendered from $24$ azimuth angles with a fixed elevation angle, under the same camera setup, and lighting setup using Blender. The render size was $64 \times 64$ pixels. We used the same training, validation, and test sets as those used in~\cite{yan2016perspective}.

% \paragraph{Baselines.}
We compared reconstruction accuracy between the voxel-based and retrieval-based approaches~\cite{yan2016perspective}. In the voxel-based approach, $G(x)$ is composed of a convolutional encoder and deconvolutional decoder. While their encoder was pre-trained using the method in Yang \etal~\cite{yang2015weakly}, our network works well without any pre-training. In the retrieval-based approach, the nearest training image is retrieved using the \cname{fc6} feature of a pre-trained VGG network~\cite{simonyan2014very}. The corresponding voxels are regarded as a predicted shape. Note that the retrieval-based approach uses ground truth voxels for supervision.

% \paragraph{Evaluation metrics.}
To evaluate the reconstruction performance quantitatively, we voxelized both the ground truth meshes and the generated meshes to compute the intersection over union (IoU) between the voxels. The size of voxels was set to $32^3$. For each object in the test set, we performed 3D reconstruction using the images from $24$ viewpoints, calculated the IoU scores, and reported the average score.

% \paragraph{Network architecture.}
We used an encoder-decoder architecture for the generator $G(x)$. Our encoder is nearly identical to that of \cite{yan2016perspective}, which encodes an input image into a $512$D vector. Our decoder is composed of three fully-connected layers. The sizes of the hidden layer are $1024$ and $2048$.

% \paragraph{Hyper parameters.}
The render size of our renderer is set to $128 \times 128$ and downsampled them to $64 \times 64$. We rendered only the silhouettes of objects without using textures and lighting. We set $\lambda_{\textrm{sl}} = 1$ and $\lambda_{\textrm{sm}} = 0.001$ in Section~
\ref{sec:experiment_reconstruction_qualitative}, and $\lambda_{\textrm{sm}} = 0$ in Section~\ref{sec:experiment_reconstruction_quantitative}.

% \paragraph{Training.}
We trained our generator using the Adam optimizer~\cite{kingma2014adam} with $\alpha=0.0001$, $\beta_1=0.9$, and $\beta_2=0.999$. The batch size was set to $64$. In each minibatch, we included silhouettes from two viewpoints per input image.

%---------------------------------------------------------
\subsubsection{Qualitative evaluation}
\label{sec:experiment_reconstruction_qualitative}

We trained $13$ models with images from each class. Figure~\ref{fig:reconstruction} presents a part of results from the test set by our mesh-based method and the voxel-based method~\cite{yan2016perspective}\footnote{We trained generators using the code from the authors and our dataset.}. Additional results are presented in the supplementary materials. These results demonstrate that a mesh can be correctly reconstructed from a single image using our method.

Compared to the voxel-based approach, the shapes reconstructed by our method are more visually appealing from the two points. One is that a mesh can represent small parts, such as airplane wings, with high resolution. The other is that there is no cubic artifacts in a mesh. Although low resolutions and artifacts may not be a problem in tasks such as picking by robots, they are disadvantageous for computer graphics, computational photography, and data augmentation.

Without using the smoothness loss, our model sometimes produces very rough surfaces. That is because the smoothness of surfaces has little effect on silhouettes. With the smoothness regularizer, the surface becomes smoother and looks more natural. Figure~\ref{fig:reconstruction2} illustrates the effectiveness of the regularizer. However, if the regularizer is used, the voxel IoU for the entire dataset becomes slightly lower.

%---------------------------------------------------------
\subsubsection{Quantitative evaluation}
\label{sec:experiment_reconstruction_quantitative}

We trained a single model using images from all classes. The reconstruction accuracy is shown in Table~\ref{table_reconstruction}. Our mesh-based approach outperforms the voxel-based approach~\cite{yan2016perspective} for $10$ out of $13$ categories. Our result is significantly better for the \cname{airplane}, \cname{chair}, \cname{display}, \cname{loudspeaker}, and \cname{sofa} categories. The basic shapes of the \cname{loudspeaker} and \cname{display} categories are simple. However, the size and position vary depending on the objects. The fact that a meshes are suitable for scaling and translation presumably contributes to the performance improvements in these categories. The variations in shapes in the \cname{airplane}, \cname{chair} and \cname{sofa} categories are also relatively small.

Our approach did not perform very well for the \cname{car}, \cname{lamp}, and \cname{table} categories. The shapes of the objects in these categories are relatively complicated, and they are difficult to be reconstructed by deforming a sphere.

%---------------------------------------------------------
\subsubsection{Limitation}
Although our reconstruction method already surpasses the voxel-based method in terms of visual appeal and voxel IoU, it has a clear disadvantage in that it cannot generate objects with various topologies. In order to overcome this limitation, it is necessary to generate the faces-to-vertices relationship $\{ \bm{f}_i \}$ dynamically. This is beyond the scope of this study, but it is an interesting direction for future research.

%-----------------------------------------------------------------
\subsection{Gradient-based 3D editing via 2D loss}

%---------------------------------------------------------
\subsubsection{Experimental settings}
\newcommand{\expnumber}[2]{{#1}\mathrm{e}{#2}}

\begin{figure*}[t]
    \begin{center}
        \includegraphics[height=0.138\textwidth,width=0.138\textwidth]{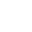}
        \includegraphics[height=0.138\textwidth,width=0.138\textwidth,trim=16 16 16 16,clip]{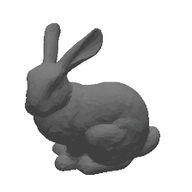}
        \includegraphics[height=0.138\textwidth,width=0.138\textwidth,trim=16 16 16 16,clip]{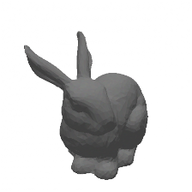}
        \includegraphics[height=0.138\textwidth,width=0.138\textwidth,trim=16 16 16 16,clip]{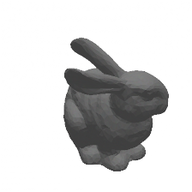}
        \includegraphics[height=0.138\textwidth,width=0.138\textwidth,trim=16 16 16 16,clip]{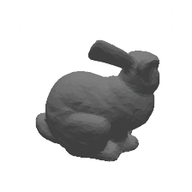}
        \includegraphics[height=0.138\textwidth,width=0.138\textwidth,trim=16 16 16 16,clip]{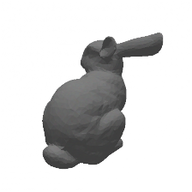}
        \includegraphics[height=0.138\textwidth,width=0.138\textwidth,trim=16 16 16 16,clip]{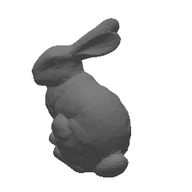} \\
        \includegraphics[height=0.138\textwidth,width=0.138\textwidth]{./figure/white.png}
        \includegraphics[height=0.138\textwidth,width=0.138\textwidth,trim=16 16 16 16,clip]{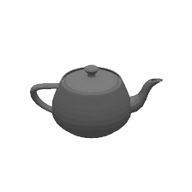}
        \includegraphics[height=0.138\textwidth,width=0.138\textwidth,trim=16 16 16 16,clip]{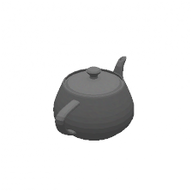}
        \includegraphics[height=0.138\textwidth,width=0.138\textwidth,trim=16 16 16 16,clip]{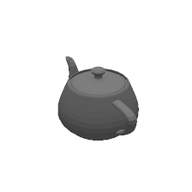}
        \includegraphics[height=0.138\textwidth,width=0.138\textwidth,trim=16 16 16 16,clip]{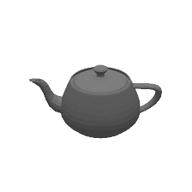}
        \includegraphics[height=0.138\textwidth,width=0.138\textwidth,trim=16 16 16 16,clip]{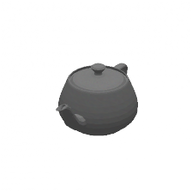}
        \includegraphics[height=0.138\textwidth,width=0.138\textwidth,trim=16 16 16 16,clip]{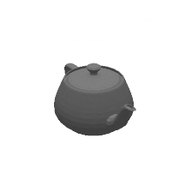} \\
    \end{center}
    \vspace{-2mm}
    \caption{Initial state of meshes in style transfer and DeepDream. Rendered from six viewpoints.}
    \label{fig:bunny_and_teapot}
\end{figure*}

We applied 2D-to-3D style transfer and 3D DeepDream to the objects shown in Figure~\ref{fig:bunny_and_teapot}. Optimization was conducted using the Adam optimizer~\cite{kingma2014adam} with $\beta_1=0.9$, and $\beta_2=0.999$. We rendered images of size $448 \times 448$ and downsampled them to $244 \times 224$ to eliminate aliasing. The batch size was set to $4$. During optimization, images were rendered at random elevations and azimuth angles. Texture size was set to $s_t = 4$.

For style transfer, the style images we used were selected from~\cite{dumoulin2016learned,johnson2016perceptual}. $\lambda_c$, $\lambda_s$, and $\lambda_t$ are manually tuned for each input. The feature extractors $f_s$ for style loss were \cname{conv1\_2}, \cname{conv2\_3}, \cname{conv3\_3}, and \cname{conv4\_3} from the VGG-16 network~\cite{simonyan2014very}. The intensities of the lights were $l^a=0.5$ and $l^d=0.5$, and the direction of the light was randomly set during optimization. The $\alpha$ value of Adam was set to $\expnumber{2.5}{-4}, \expnumber{5}{-2}$ for $\{ \bm{v}_i \}, \{ \bm{t}_i \}$. The number of parameter updates was set to $5,000$.

In DeepDream, images are rendered without lighting. The feature extractor was the \cname{inception\_4c} layer from GoogLeNet~\cite{szegedy2015going}. The $\alpha$ value of Adam was set to $\expnumber{5}{-5}, \expnumber{1}{-2}$ for $\{ \bm{v}_i \}, \{ \bm{t}_i \}$. Optimization is stopped after $1,000$ iterations.

%---------------------------------------------------------
\subsubsection{2D-to-3D Style Transfer}
Figure~\ref{fig:style_transfer} presents the results of 2D-to-3D style transfer. Additional results are shown in the supplementary materials.

The styles of the paintings were accurately transferred to the textures and shapes. From the outline of the bunny and the lid of the teapot, we can see the straight style of Coupland and Gris. The wavy style of Munch was also transferred to the side of the teapot. Interestingly, the side of the tower of Babel was transferred only to the side, not to the upside, of the bunny.

The proposed method provides a way to edit 3D models intuitively and quickly. This can be useful for rapid prototyping for product design as well as art production.

%---------------------------------------------------------
\subsubsection{3D DeepDream}
\begin{figure*}[t]
    \begin{center}
        \includegraphics[height=0.138\textwidth,width=0.138\textwidth]{./figure/white.png}
        \includegraphics[height=0.138\textwidth,width=0.138\textwidth,trim=16 16 16 16,clip]{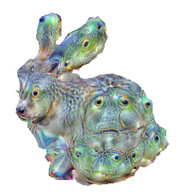}
        \includegraphics[height=0.138\textwidth,width=0.138\textwidth,trim=16 16 16 16,clip]{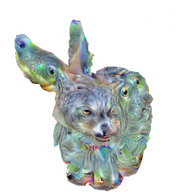}
        \includegraphics[height=0.138\textwidth,width=0.138\textwidth,trim=16 16 16 16,clip]{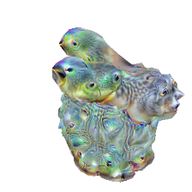}
        \includegraphics[height=0.138\textwidth,width=0.138\textwidth,trim=16 16 16 16,clip]{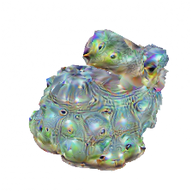}
        \includegraphics[height=0.138\textwidth,width=0.138\textwidth,trim=16 16 16 16,clip]{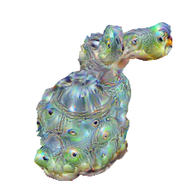}
        \includegraphics[height=0.138\textwidth,width=0.138\textwidth,trim=16 16 16 16,clip]{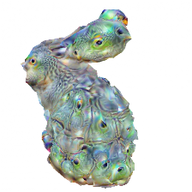} \\
        \includegraphics[height=0.138\textwidth,width=0.138\textwidth]{./figure/white.png}
        \includegraphics[height=0.138\textwidth,width=0.138\textwidth,trim=16 16 16 16,clip]{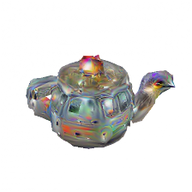}
        \includegraphics[height=0.138\textwidth,width=0.138\textwidth,trim=16 16 16 16,clip]{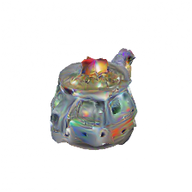}
        \includegraphics[height=0.138\textwidth,width=0.138\textwidth,trim=16 16 16 16,clip]{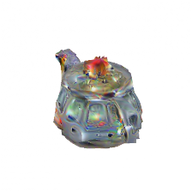}
        \includegraphics[height=0.138\textwidth,width=0.138\textwidth,trim=16 16 16 16,clip]{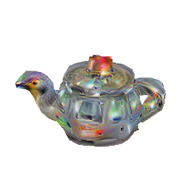}
        \includegraphics[height=0.138\textwidth,width=0.138\textwidth,trim=16 16 16 16,clip]{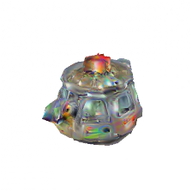}
        \includegraphics[height=0.138\textwidth,width=0.138\textwidth,trim=16 16 16 16,clip]{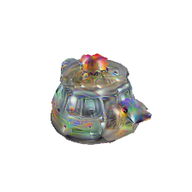} \\
    \end{center}
    \vspace{-2mm}
    \caption{DeepDream of 3D mesh.}
    \label{fig:deep_dream}
\end{figure*}

Figure~\ref{fig:deep_dream} presents the results of DeepDream. A nose and eyes emerged on the face of the bunny. The spout of the teapot expanded and became the face of the bird, while the body appeared similar to a bus. These transformations matched the 3D shape of each object.

%-------------------------------------------------------------------------
\section{Conclusion}

In this paper, we enabled the integration of rendering of a 3D mesh into neural networks by proposing an approximate gradient for rendering. Using this renderer, we proposed a method to reconstruct a 3D mesh from a single image, the performance of which is superior to the existing voxel-based approach~\cite{yan2016perspective} in terms of visual appeal and the voxel IoU metric. We also proposed a method to edit the vertices and textures of a 3D mesh according to its 3D shape using a loss function on images and gradient descent. These applications demonstrate the potential of integrating mesh renderers into neural networks and the effectiveness of the proposed renderer.

The applications of our renderer are not limited to those presented in this paper. Other problems will be solved through incorporating our module in other systems.

%-------------------------------------------------------------------------
\clearpage

\begin{figure*}[t]
    \begin{center}
        \includegraphics[height=0.138\textwidth,width=0.138\textwidth]{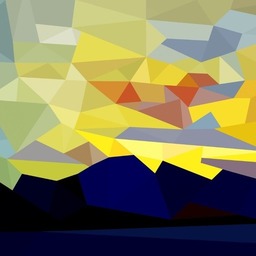}
        \includegraphics[height=0.138\textwidth,width=0.138\textwidth,trim=16 16 16 16,clip]{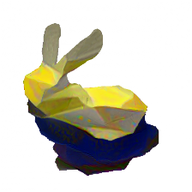}
        \includegraphics[height=0.138\textwidth,width=0.138\textwidth,trim=16 16 16 16,clip]{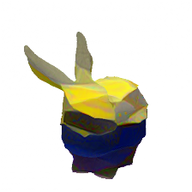}
        \includegraphics[height=0.138\textwidth,width=0.138\textwidth,trim=16 16 16 16,clip]{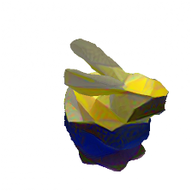}
        \includegraphics[height=0.138\textwidth,width=0.138\textwidth,trim=16 16 16 16,clip]{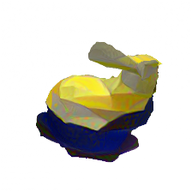}
        \includegraphics[height=0.138\textwidth,width=0.138\textwidth,trim=16 16 16 16,clip]{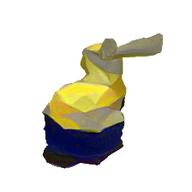}
        \includegraphics[height=0.138\textwidth,width=0.138\textwidth,trim=16 16 16 16,clip]{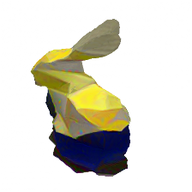} \\
        \includegraphics[height=0.138\textwidth,width=0.138\textwidth]{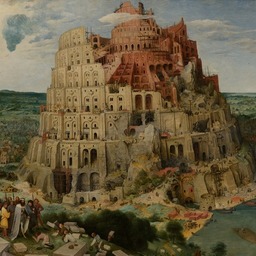}
        \includegraphics[height=0.138\textwidth,width=0.138\textwidth,trim=16 16 16 16,clip]{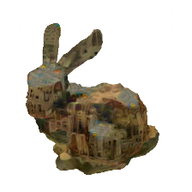}
        \includegraphics[height=0.138\textwidth,width=0.138\textwidth,trim=16 16 16 16,clip]{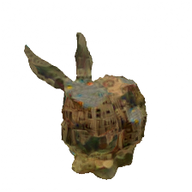}
        \includegraphics[height=0.138\textwidth,width=0.138\textwidth,trim=16 16 16 16,clip]{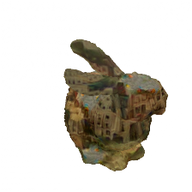}
        \includegraphics[height=0.138\textwidth,width=0.138\textwidth,trim=16 16 16 16,clip]{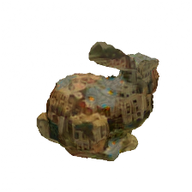}
        \includegraphics[height=0.138\textwidth,width=0.138\textwidth,trim=16 16 16 16,clip]{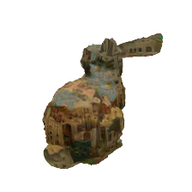}
        \includegraphics[height=0.138\textwidth,width=0.138\textwidth,trim=16 16 16 16,clip]{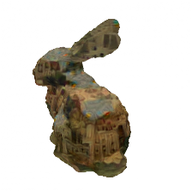} \\
        \includegraphics[height=0.138\textwidth,width=0.138\textwidth]{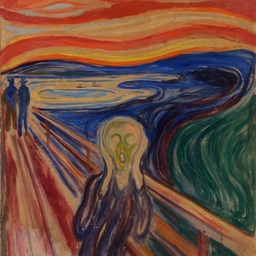}
        \includegraphics[height=0.138\textwidth,width=0.138\textwidth,trim=16 16 16 16,clip]{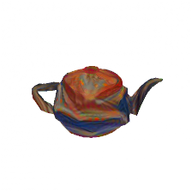}
        \includegraphics[height=0.138\textwidth,width=0.138\textwidth,trim=16 16 16 16,clip]{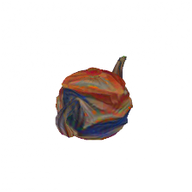}
        \includegraphics[height=0.138\textwidth,width=0.138\textwidth,trim=16 16 16 16,clip]{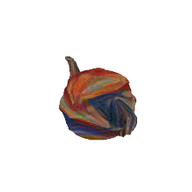}
        \includegraphics[height=0.138\textwidth,width=0.138\textwidth,trim=16 16 16 16,clip]{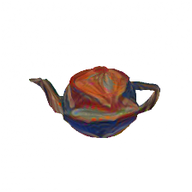}
        \includegraphics[height=0.138\textwidth,width=0.138\textwidth,trim=16 16 16 16,clip]{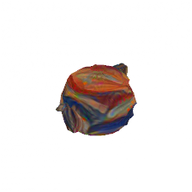}
        \includegraphics[height=0.138\textwidth,width=0.138\textwidth,trim=16 16 16 16,clip]{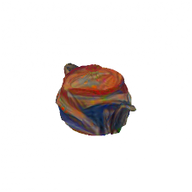} \\
        \includegraphics[height=0.138\textwidth,width=0.138\textwidth]{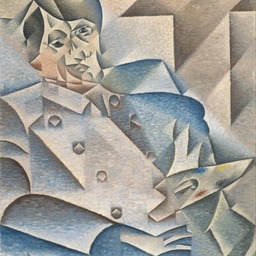}
        \includegraphics[height=0.138\textwidth,width=0.138\textwidth,trim=16 16 16 16,clip]{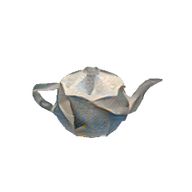}
        \includegraphics[height=0.138\textwidth,width=0.138\textwidth,trim=16 16 16 16,clip]{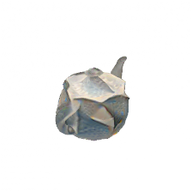}
        \includegraphics[height=0.138\textwidth,width=0.138\textwidth,trim=16 16 16 16,clip]{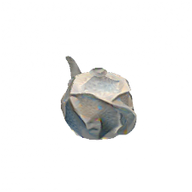}
        \includegraphics[height=0.138\textwidth,width=0.138\textwidth,trim=16 16 16 16,clip]{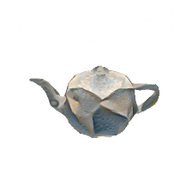}
        \includegraphics[height=0.138\textwidth,width=0.138\textwidth,trim=16 16 16 16,clip]{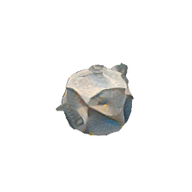}
        \includegraphics[height=0.138\textwidth,width=0.138\textwidth,trim=16 16 16 16,clip]{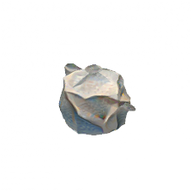} \\
    \end{center}
    \vspace{-2mm}
    \caption{2D-to-3D style transfer. The leftmost images represent styles. The style images are {\it Thomson No. 5 (Yellow Sunset)} (D. Coupland, 2011), {\it The Tower of Babel} (P. Bruegel the Elder, 1563), {\it The Scream} (E. Munch, 1910), and {\it Portrait of Pablo Picasso} (J. Gris, 1912).}
    \label{fig:style_transfer}
\end{figure*}

%-------------------------------------------------------------------------
\clearpage
\section*{Acknowledgment}
\addcontentsline{toc}{section}{Acknowledgment}
This work was partially funded by ImPACT Program of Council for Science, Technology and Innovation (Cabinet Office, Government of Japan) and partially supported by JST CREST Grant Number JPMJCR1403, Japan.

%-------------------------------------------------------------------------
{
    \small
    \bibliographystyle{ieee}
    \bibliography{paper}
}

% -------------------------------------------------------------------------
\clearpage
\appendix
\def\thesection{Appendix \Alph{section}}

% -------------------------------------------------------------------------
\section{Additional results}
\label{sec:appendix}
\begin{figure*}[t]
    \begin{center}
        \includegraphics[height=0.138\textwidth,width=0.138\textwidth]{./images/original_02691156_d6ca5966c5ed5b86da2b0f839aba40f9_45.png}
        \includegraphics[height=0.138\textwidth,width=0.138\textwidth]{./images/ours_02691156_d6ca5966c5ed5b86da2b0f839aba40f9_45_000.png}
        \includegraphics[height=0.138\textwidth,width=0.138\textwidth]{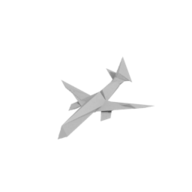}
        \includegraphics[height=0.138\textwidth,width=0.138\textwidth]{./images/ours_02691156_d6ca5966c5ed5b86da2b0f839aba40f9_45_120.png}
        \includegraphics[height=0.138\textwidth,width=0.138\textwidth]{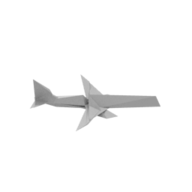}
        \includegraphics[height=0.138\textwidth,width=0.138\textwidth]{./images/ours_02691156_d6ca5966c5ed5b86da2b0f839aba40f9_45_240.png}
        \includegraphics[height=0.138\textwidth,width=0.138\textwidth]{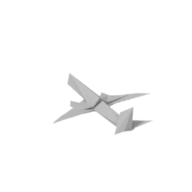} \\
        \includegraphics[height=0.138\textwidth,width=0.138\textwidth]{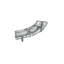}
        \includegraphics[height=0.138\textwidth,width=0.138\textwidth]{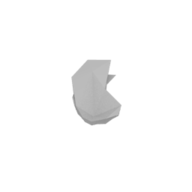}
        \includegraphics[height=0.138\textwidth,width=0.138\textwidth]{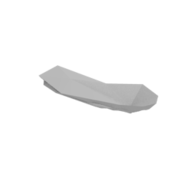}
        \includegraphics[height=0.138\textwidth,width=0.138\textwidth]{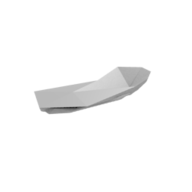}
        \includegraphics[height=0.138\textwidth,width=0.138\textwidth]{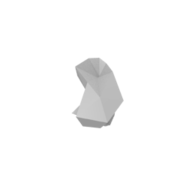}
        \includegraphics[height=0.138\textwidth,width=0.138\textwidth]{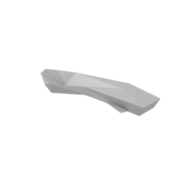}
        \includegraphics[height=0.138\textwidth,width=0.138\textwidth]{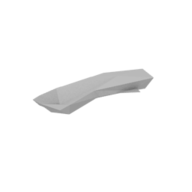} \\
        \includegraphics[height=0.138\textwidth,width=0.138\textwidth]{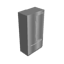}
        \includegraphics[height=0.138\textwidth,width=0.138\textwidth]{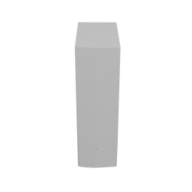}
        \includegraphics[height=0.138\textwidth,width=0.138\textwidth]{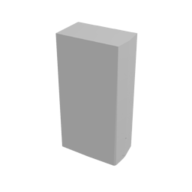}
        \includegraphics[height=0.138\textwidth,width=0.138\textwidth]{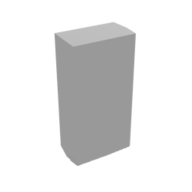}
        \includegraphics[height=0.138\textwidth,width=0.138\textwidth]{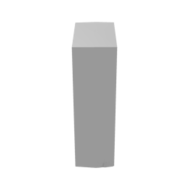}
        \includegraphics[height=0.138\textwidth,width=0.138\textwidth]{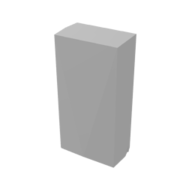}
        \includegraphics[height=0.138\textwidth,width=0.138\textwidth]{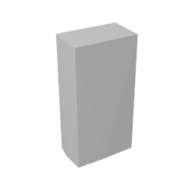} \\
        \includegraphics[height=0.138\textwidth,width=0.138\textwidth]{./images/original_02958343_e64dd4ff16eab5752a9eb0f146e94477_45.png}
        \includegraphics[height=0.138\textwidth,width=0.138\textwidth]{./images/ours_02958343_e64dd4ff16eab5752a9eb0f146e94477_45_000.png}
        \includegraphics[height=0.138\textwidth,width=0.138\textwidth]{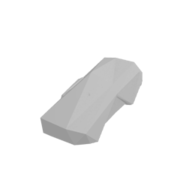}
        \includegraphics[height=0.138\textwidth,width=0.138\textwidth]{./images/ours_02958343_e64dd4ff16eab5752a9eb0f146e94477_45_120.png}
        \includegraphics[height=0.138\textwidth,width=0.138\textwidth]{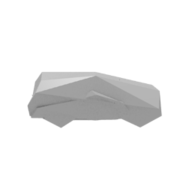}
        \includegraphics[height=0.138\textwidth,width=0.138\textwidth]{./images/ours_02958343_e64dd4ff16eab5752a9eb0f146e94477_45_240.png}
        \includegraphics[height=0.138\textwidth,width=0.138\textwidth]{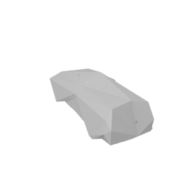} \\
        \includegraphics[height=0.138\textwidth,width=0.138\textwidth]{./images/original_03001627_9b359e42a5bc98572085b87de8f7581b_45.png}
        \includegraphics[height=0.138\textwidth,width=0.138\textwidth]{./images/ours_03001627_9b359e42a5bc98572085b87de8f7581b_45_000.png}
        \includegraphics[height=0.138\textwidth,width=0.138\textwidth]{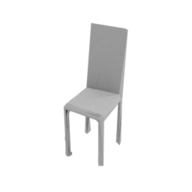}
        \includegraphics[height=0.138\textwidth,width=0.138\textwidth]{./images/ours_03001627_9b359e42a5bc98572085b87de8f7581b_45_120.png}
        \includegraphics[height=0.138\textwidth,width=0.138\textwidth]{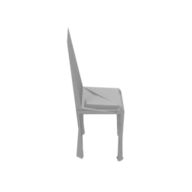}
        \includegraphics[height=0.138\textwidth,width=0.138\textwidth]{./images/ours_03001627_9b359e42a5bc98572085b87de8f7581b_45_240.png}
        \includegraphics[height=0.138\textwidth,width=0.138\textwidth]{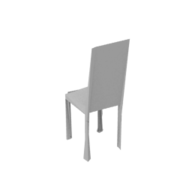} \\
        \includegraphics[height=0.138\textwidth,width=0.138\textwidth]{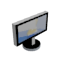}
        \includegraphics[height=0.138\textwidth,width=0.138\textwidth]{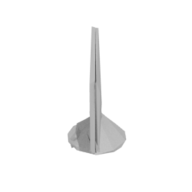}
        \includegraphics[height=0.138\textwidth,width=0.138\textwidth]{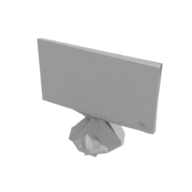}
        \includegraphics[height=0.138\textwidth,width=0.138\textwidth]{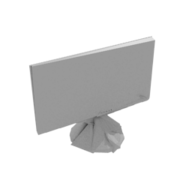}
        \includegraphics[height=0.138\textwidth,width=0.138\textwidth]{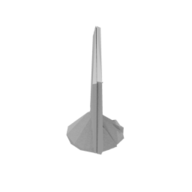}
        \includegraphics[height=0.138\textwidth,width=0.138\textwidth]{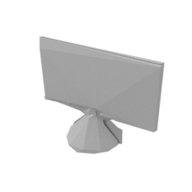}
        \includegraphics[height=0.138\textwidth,width=0.138\textwidth]{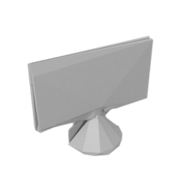} \\
        \includegraphics[height=0.138\textwidth,width=0.138\textwidth]{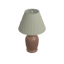}
        \includegraphics[height=0.138\textwidth,width=0.138\textwidth]{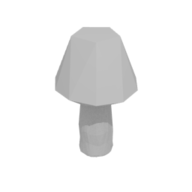}
        \includegraphics[height=0.138\textwidth,width=0.138\textwidth]{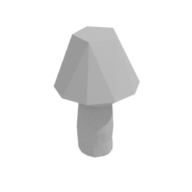}
        \includegraphics[height=0.138\textwidth,width=0.138\textwidth]{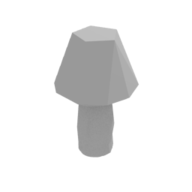}
        \includegraphics[height=0.138\textwidth,width=0.138\textwidth]{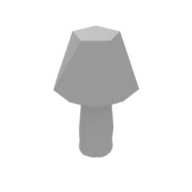}
        \includegraphics[height=0.138\textwidth,width=0.138\textwidth]{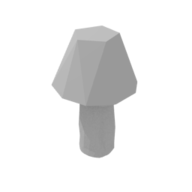}
        \includegraphics[height=0.138\textwidth,width=0.138\textwidth]{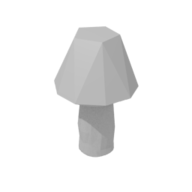} \\
    \end{center}
    \caption{3D mesh reconstruction from a single image. The leftmost images are the inputs. Results are rendered from six viewpoints.}
    \label{fig:appendix_reconstruction1}
\end{figure*}

\begin{figure*}[t]
    \begin{center}
        \includegraphics[height=0.138\textwidth,width=0.138\textwidth]{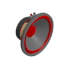}
        \includegraphics[height=0.138\textwidth,width=0.138\textwidth]{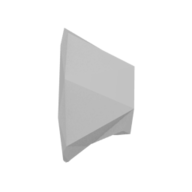}
        \includegraphics[height=0.138\textwidth,width=0.138\textwidth]{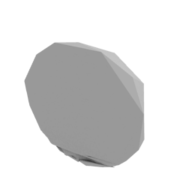}
        \includegraphics[height=0.138\textwidth,width=0.138\textwidth]{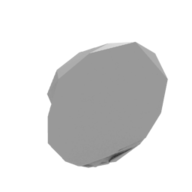}
        \includegraphics[height=0.138\textwidth,width=0.138\textwidth]{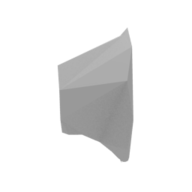}
        \includegraphics[height=0.138\textwidth,width=0.138\textwidth]{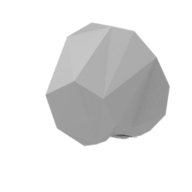}
        \includegraphics[height=0.138\textwidth,width=0.138\textwidth]{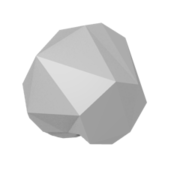} \\
        \includegraphics[height=0.138\textwidth,width=0.138\textwidth]{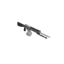}
        \includegraphics[height=0.138\textwidth,width=0.138\textwidth]{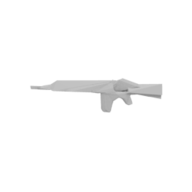}
        \includegraphics[height=0.138\textwidth,width=0.138\textwidth]{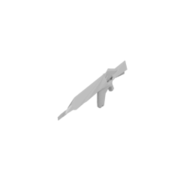}
        \includegraphics[height=0.138\textwidth,width=0.138\textwidth]{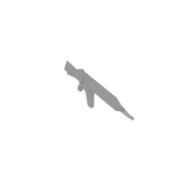}
        \includegraphics[height=0.138\textwidth,width=0.138\textwidth]{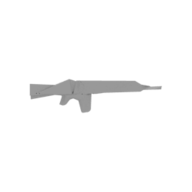}
        \includegraphics[height=0.138\textwidth,width=0.138\textwidth]{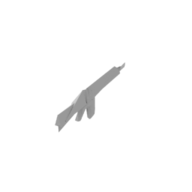}
        \includegraphics[height=0.138\textwidth,width=0.138\textwidth]{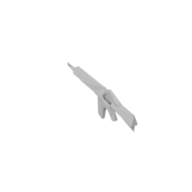} \\
        \includegraphics[height=0.138\textwidth,width=0.138\textwidth]{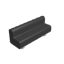}
        \includegraphics[height=0.138\textwidth,width=0.138\textwidth]{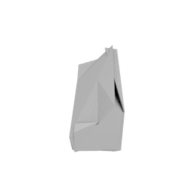}
        \includegraphics[height=0.138\textwidth,width=0.138\textwidth]{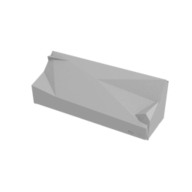}
        \includegraphics[height=0.138\textwidth,width=0.138\textwidth]{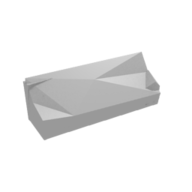}
        \includegraphics[height=0.138\textwidth,width=0.138\textwidth]{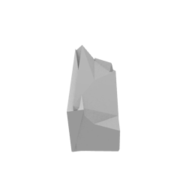}
        \includegraphics[height=0.138\textwidth,width=0.138\textwidth]{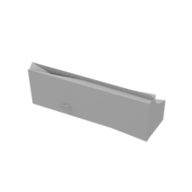}
        \includegraphics[height=0.138\textwidth,width=0.138\textwidth]{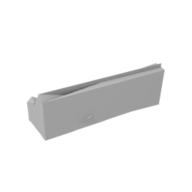} \\
        \includegraphics[height=0.138\textwidth,width=0.138\textwidth]{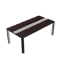}
        \includegraphics[height=0.138\textwidth,width=0.138\textwidth]{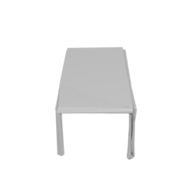}
        \includegraphics[height=0.138\textwidth,width=0.138\textwidth]{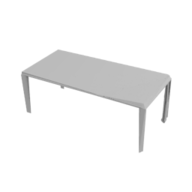}
        \includegraphics[height=0.138\textwidth,width=0.138\textwidth]{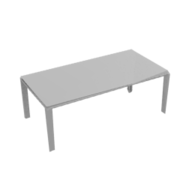}
        \includegraphics[height=0.138\textwidth,width=0.138\textwidth]{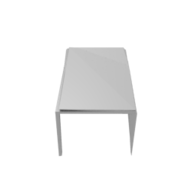}
        \includegraphics[height=0.138\textwidth,width=0.138\textwidth]{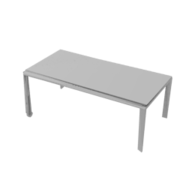}
        \includegraphics[height=0.138\textwidth,width=0.138\textwidth]{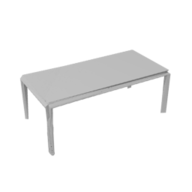} \\
        \includegraphics[height=0.138\textwidth,width=0.138\textwidth]{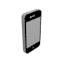}
        \includegraphics[height=0.138\textwidth,width=0.138\textwidth]{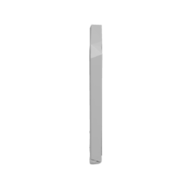}
        \includegraphics[height=0.138\textwidth,width=0.138\textwidth]{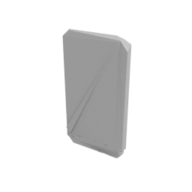}
        \includegraphics[height=0.138\textwidth,width=0.138\textwidth]{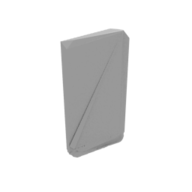}
        \includegraphics[height=0.138\textwidth,width=0.138\textwidth]{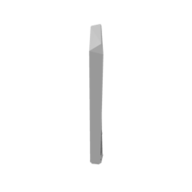}
        \includegraphics[height=0.138\textwidth,width=0.138\textwidth]{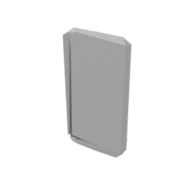}
        \includegraphics[height=0.138\textwidth,width=0.138\textwidth]{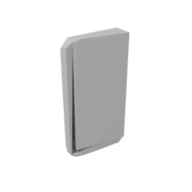} \\
        \includegraphics[height=0.138\textwidth,width=0.138\textwidth]{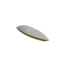}
        \includegraphics[height=0.138\textwidth,width=0.138\textwidth]{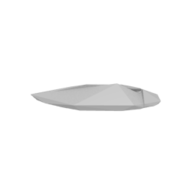}
        \includegraphics[height=0.138\textwidth,width=0.138\textwidth]{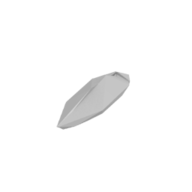}
        \includegraphics[height=0.138\textwidth,width=0.138\textwidth]{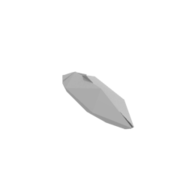}
        \includegraphics[height=0.138\textwidth,width=0.138\textwidth]{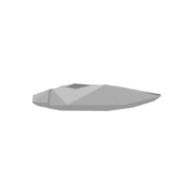}
        \includegraphics[height=0.138\textwidth,width=0.138\textwidth]{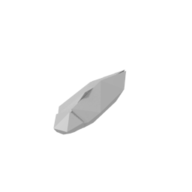}
        \includegraphics[height=0.138\textwidth,width=0.138\textwidth]{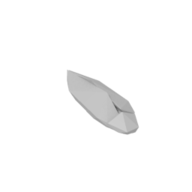} \\
    \end{center}
    \caption{3D mesh reconstruction from a single image. The leftmost images are the inputs. Results are rendered from six viewpoints.}
    \label{fig:appendix_reconstruction2}
\end{figure*}

\begin{figure*}[t]
    \begin{center}
        \includegraphics[height=0.138\textwidth,width=0.138\textwidth]{./figure/white.png}
        \includegraphics[height=0.138\textwidth,width=0.138\textwidth]{./images/bunny/rotation_00000000.png}
        \includegraphics[height=0.138\textwidth,width=0.138\textwidth]{./images/bunny/rotation_00000300.png}
        \includegraphics[height=0.138\textwidth,width=0.138\textwidth]{./images/bunny/rotation_00000240.png}
        \includegraphics[height=0.138\textwidth,width=0.138\textwidth]{./images/bunny/rotation_00000180.png}
        \includegraphics[height=0.138\textwidth,width=0.138\textwidth]{./images/bunny/rotation_00000120.png}
        \includegraphics[height=0.138\textwidth,width=0.138\textwidth]{./images/bunny/rotation_00000060.png} \\
        \includegraphics[height=0.138\textwidth,width=0.138\textwidth]{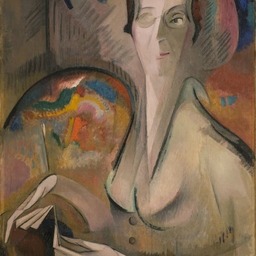}
        \includegraphics[height=0.138\textwidth,width=0.138\textwidth]{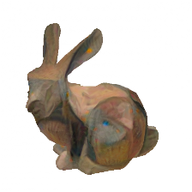}
        \includegraphics[height=0.138\textwidth,width=0.138\textwidth]{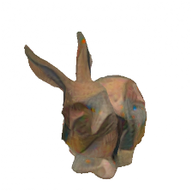}
        \includegraphics[height=0.138\textwidth,width=0.138\textwidth]{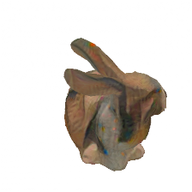}
        \includegraphics[height=0.138\textwidth,width=0.138\textwidth]{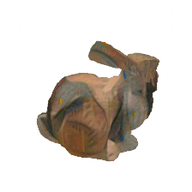}
        \includegraphics[height=0.138\textwidth,width=0.138\textwidth]{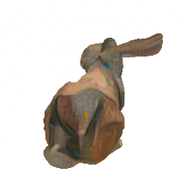}
        \includegraphics[height=0.138\textwidth,width=0.138\textwidth]{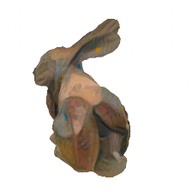} \\
        \includegraphics[height=0.138\textwidth,width=0.138\textwidth]{./images/style_transfer/coupland1.jpg}
        \includegraphics[height=0.138\textwidth,width=0.138\textwidth]{./images/bunny_coupland1_lc_10000000000_ltv_100000/rotation_00000000.png}
        \includegraphics[height=0.138\textwidth,width=0.138\textwidth]{./images/bunny_coupland1_lc_10000000000_ltv_100000/rotation_00000300.png}
        \includegraphics[height=0.138\textwidth,width=0.138\textwidth]{./images/bunny_coupland1_lc_10000000000_ltv_100000/rotation_00000240.png}
        \includegraphics[height=0.138\textwidth,width=0.138\textwidth]{./images/bunny_coupland1_lc_10000000000_ltv_100000/rotation_00000180.png}
        \includegraphics[height=0.138\textwidth,width=0.138\textwidth]{./images/bunny_coupland1_lc_10000000000_ltv_100000/rotation_00000120.png}
        \includegraphics[height=0.138\textwidth,width=0.138\textwidth]{./images/bunny_coupland1_lc_10000000000_ltv_100000/rotation_00000060.png} \\
        \includegraphics[height=0.138\textwidth,width=0.138\textwidth]{./images/style_transfer/elder1.jpg}
        \includegraphics[height=0.138\textwidth,width=0.138\textwidth]{./images/bunny_elder1_lc_1000000000_ltv_100000/rotation_00000000.png}
        \includegraphics[height=0.138\textwidth,width=0.138\textwidth]{./images/bunny_elder1_lc_1000000000_ltv_100000/rotation_00000300.png}
        \includegraphics[height=0.138\textwidth,width=0.138\textwidth]{./images/bunny_elder1_lc_1000000000_ltv_100000/rotation_00000240.png}
        \includegraphics[height=0.138\textwidth,width=0.138\textwidth]{./images/bunny_elder1_lc_1000000000_ltv_100000/rotation_00000180.png}
        \includegraphics[height=0.138\textwidth,width=0.138\textwidth]{./images/bunny_elder1_lc_1000000000_ltv_100000/rotation_00000120.png}
        \includegraphics[height=0.138\textwidth,width=0.138\textwidth]{./images/bunny_elder1_lc_1000000000_ltv_100000/rotation_00000060.png} \\
        \includegraphics[height=0.138\textwidth,width=0.138\textwidth]{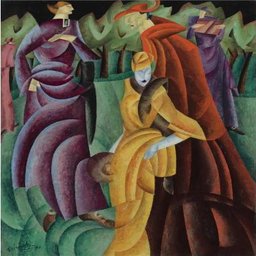}
        \includegraphics[height=0.138\textwidth,width=0.138\textwidth]{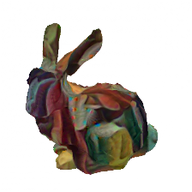}
        \includegraphics[height=0.138\textwidth,width=0.138\textwidth]{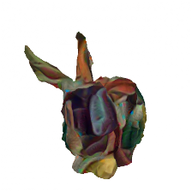}
        \includegraphics[height=0.138\textwidth,width=0.138\textwidth]{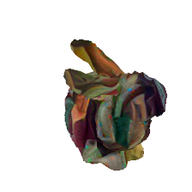}
        \includegraphics[height=0.138\textwidth,width=0.138\textwidth]{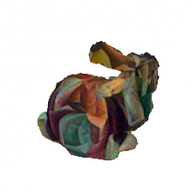}
        \includegraphics[height=0.138\textwidth,width=0.138\textwidth]{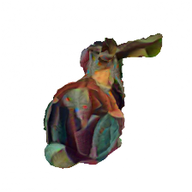}
        \includegraphics[height=0.138\textwidth,width=0.138\textwidth]{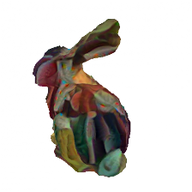} \\
        \includegraphics[height=0.138\textwidth,width=0.138\textwidth]{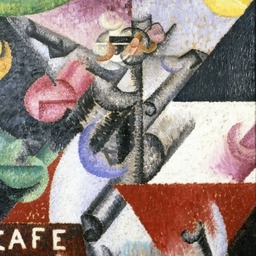}
        \includegraphics[height=0.138\textwidth,width=0.138\textwidth]{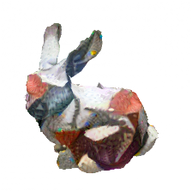}
        \includegraphics[height=0.138\textwidth,width=0.138\textwidth]{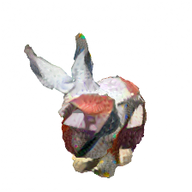}
        \includegraphics[height=0.138\textwidth,width=0.138\textwidth]{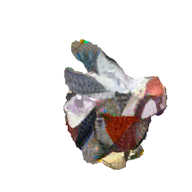}
        \includegraphics[height=0.138\textwidth,width=0.138\textwidth]{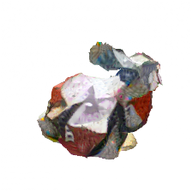}
        \includegraphics[height=0.138\textwidth,width=0.138\textwidth]{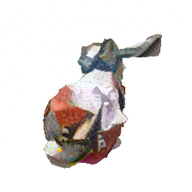}
        \includegraphics[height=0.138\textwidth,width=0.138\textwidth]{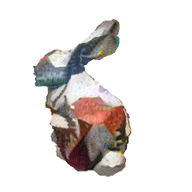} \\
        \includegraphics[height=0.138\textwidth,width=0.138\textwidth]{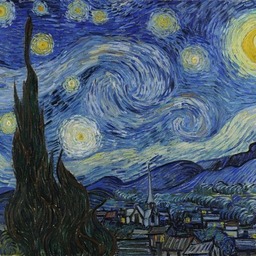}
        \includegraphics[height=0.138\textwidth,width=0.138\textwidth]{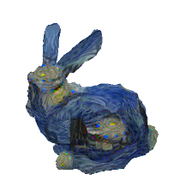}
        \includegraphics[height=0.138\textwidth,width=0.138\textwidth]{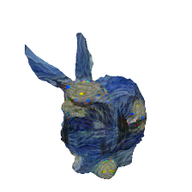}
        \includegraphics[height=0.138\textwidth,width=0.138\textwidth]{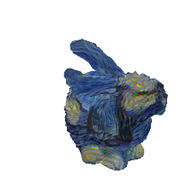}
        \includegraphics[height=0.138\textwidth,width=0.138\textwidth]{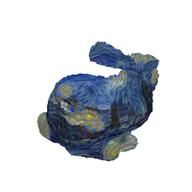}
        \includegraphics[height=0.138\textwidth,width=0.138\textwidth]{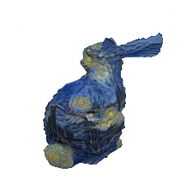}
        \includegraphics[height=0.138\textwidth,width=0.138\textwidth]{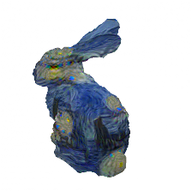} \\
        \includegraphics[height=0.138\textwidth,width=0.138\textwidth]{./images/style_transfer/gris1.jpg}
        \includegraphics[height=0.138\textwidth,width=0.138\textwidth]{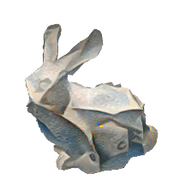}
        \includegraphics[height=0.138\textwidth,width=0.138\textwidth]{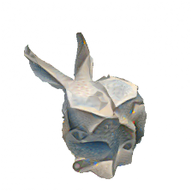}
        \includegraphics[height=0.138\textwidth,width=0.138\textwidth]{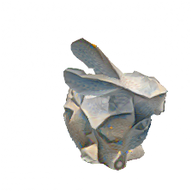}
        \includegraphics[height=0.138\textwidth,width=0.138\textwidth]{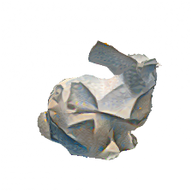}
        \includegraphics[height=0.138\textwidth,width=0.138\textwidth]{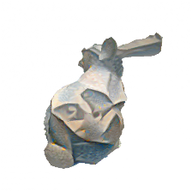}
        \includegraphics[height=0.138\textwidth,width=0.138\textwidth]{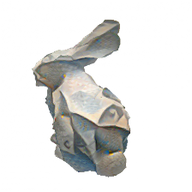} \\
    \end{center}
    \caption{Additional results of style transfer. The style images are {\it Self-Portrait} (A. Bailly, 1917), {\it Thomson No. 5 (Yellow Sunset)} (D. Coupland, 2011), {\it The Tower of Babel} (P. Bruegel the Elder, 1563), {\it Jesuits III} (L. Feininger, 1915), {\it Ritmo plastico del 14 luglio} (S. Gino, 1913), {\it The Starry Night} (V. van Gogh, 1889), and {\it Portrait of Pablo Picasso} (J. Gris, 1912).}
    \label{fig:appendix_style_transfer1}
\end{figure*}

\begin{figure*}[t]
    \begin{center}
        \includegraphics[height=0.138\textwidth,width=0.138\textwidth]{./figure/white.png}
        \includegraphics[height=0.138\textwidth,width=0.138\textwidth]{./images/bunny/rotation_00000000.png}
        \includegraphics[height=0.138\textwidth,width=0.138\textwidth]{./images/bunny/rotation_00000300.png}
        \includegraphics[height=0.138\textwidth,width=0.138\textwidth]{./images/bunny/rotation_00000240.png}
        \includegraphics[height=0.138\textwidth,width=0.138\textwidth]{./images/bunny/rotation_00000180.png}
        \includegraphics[height=0.138\textwidth,width=0.138\textwidth]{./images/bunny/rotation_00000120.png}
        \includegraphics[height=0.138\textwidth,width=0.138\textwidth]{./images/bunny/rotation_00000060.png} \\
        \includegraphics[height=0.138\textwidth,width=0.138\textwidth]{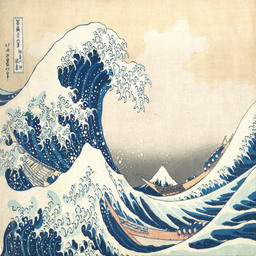}
        \includegraphics[height=0.138\textwidth,width=0.138\textwidth]{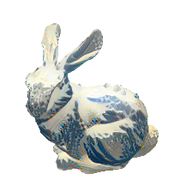}
        \includegraphics[height=0.138\textwidth,width=0.138\textwidth]{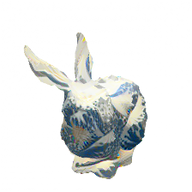}
        \includegraphics[height=0.138\textwidth,width=0.138\textwidth]{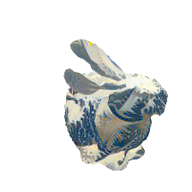}
        \includegraphics[height=0.138\textwidth,width=0.138\textwidth]{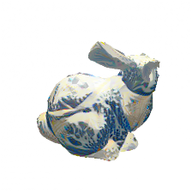}
        \includegraphics[height=0.138\textwidth,width=0.138\textwidth]{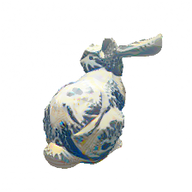}
        \includegraphics[height=0.138\textwidth,width=0.138\textwidth]{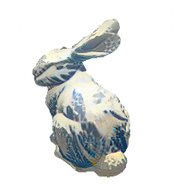} \\
        \includegraphics[height=0.138\textwidth,width=0.138\textwidth]{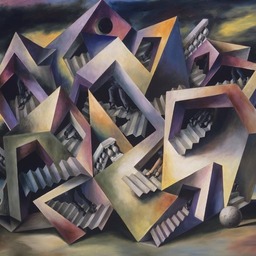}
        \includegraphics[height=0.138\textwidth,width=0.138\textwidth]{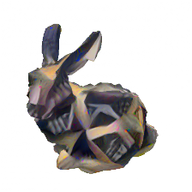}
        \includegraphics[height=0.138\textwidth,width=0.138\textwidth]{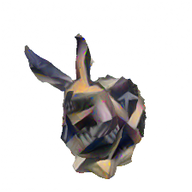}
        \includegraphics[height=0.138\textwidth,width=0.138\textwidth]{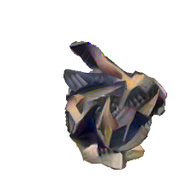}
        \includegraphics[height=0.138\textwidth,width=0.138\textwidth]{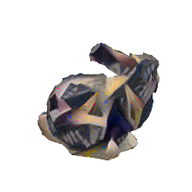}
        \includegraphics[height=0.138\textwidth,width=0.138\textwidth]{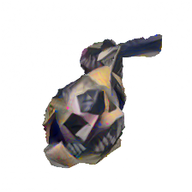}
        \includegraphics[height=0.138\textwidth,width=0.138\textwidth]{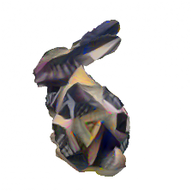} \\
        \includegraphics[height=0.138\textwidth,width=0.138\textwidth]{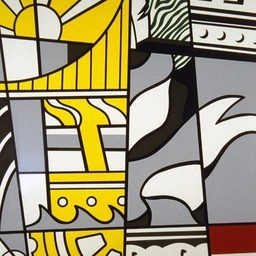}
        \includegraphics[height=0.138\textwidth,width=0.138\textwidth]{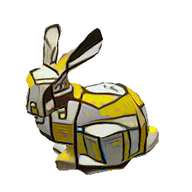}
        \includegraphics[height=0.138\textwidth,width=0.138\textwidth]{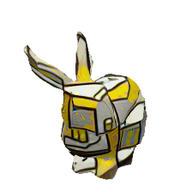}
        \includegraphics[height=0.138\textwidth,width=0.138\textwidth]{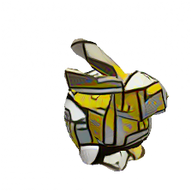}
        \includegraphics[height=0.138\textwidth,width=0.138\textwidth]{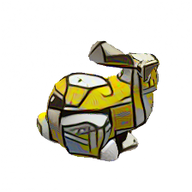}
        \includegraphics[height=0.138\textwidth,width=0.138\textwidth]{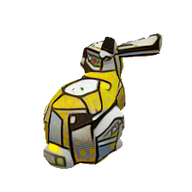}
        \includegraphics[height=0.138\textwidth,width=0.138\textwidth]{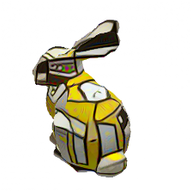} \\
        \includegraphics[height=0.138\textwidth,width=0.138\textwidth]{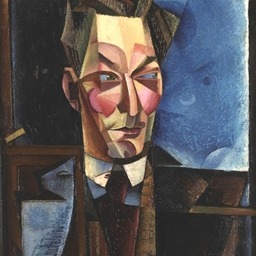}
        \includegraphics[height=0.138\textwidth,width=0.138\textwidth]{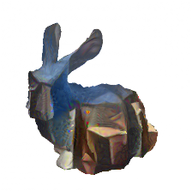}
        \includegraphics[height=0.138\textwidth,width=0.138\textwidth]{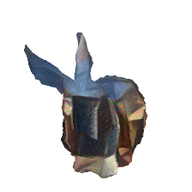}
        \includegraphics[height=0.138\textwidth,width=0.138\textwidth]{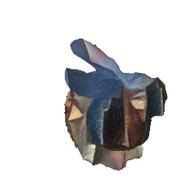}
        \includegraphics[height=0.138\textwidth,width=0.138\textwidth]{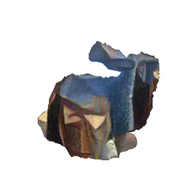}
        \includegraphics[height=0.138\textwidth,width=0.138\textwidth]{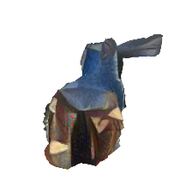}
        \includegraphics[height=0.138\textwidth,width=0.138\textwidth]{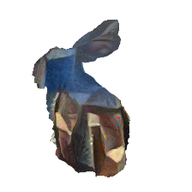} \\
        \includegraphics[height=0.138\textwidth,width=0.138\textwidth]{./images/style_transfer/munch1.jpg}
        \includegraphics[height=0.138\textwidth,width=0.138\textwidth]{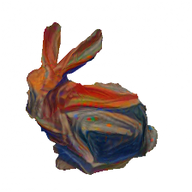}
        \includegraphics[height=0.138\textwidth,width=0.138\textwidth]{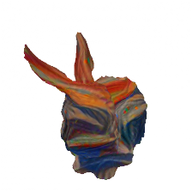}
        \includegraphics[height=0.138\textwidth,width=0.138\textwidth]{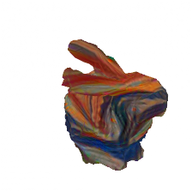}
        \includegraphics[height=0.138\textwidth,width=0.138\textwidth]{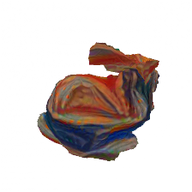}
        \includegraphics[height=0.138\textwidth,width=0.138\textwidth]{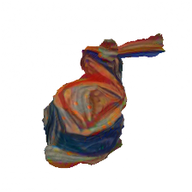}
        \includegraphics[height=0.138\textwidth,width=0.138\textwidth]{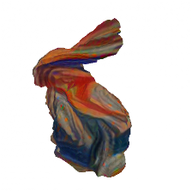} \\
        \includegraphics[height=0.138\textwidth,width=0.138\textwidth]{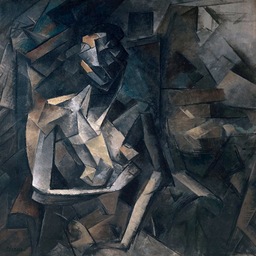}
        \includegraphics[height=0.138\textwidth,width=0.138\textwidth]{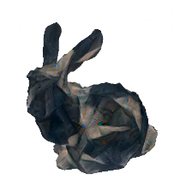}
        \includegraphics[height=0.138\textwidth,width=0.138\textwidth]{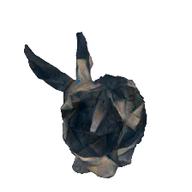}
        \includegraphics[height=0.138\textwidth,width=0.138\textwidth]{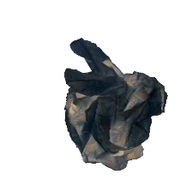}
        \includegraphics[height=0.138\textwidth,width=0.138\textwidth]{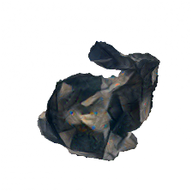}
        \includegraphics[height=0.138\textwidth,width=0.138\textwidth]{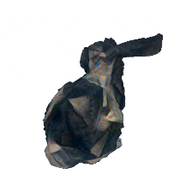}
        \includegraphics[height=0.138\textwidth,width=0.138\textwidth]{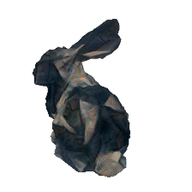} \\
        \includegraphics[height=0.138\textwidth,width=0.138\textwidth]{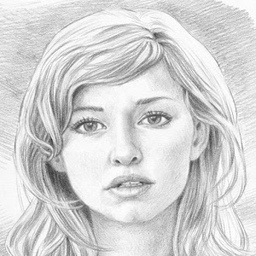}
        \includegraphics[height=0.138\textwidth,width=0.138\textwidth]{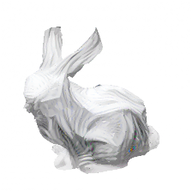}
        \includegraphics[height=0.138\textwidth,width=0.138\textwidth]{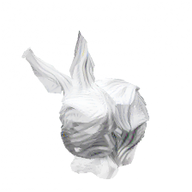}
        \includegraphics[height=0.138\textwidth,width=0.138\textwidth]{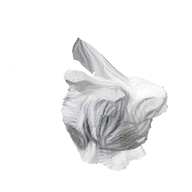}
        \includegraphics[height=0.138\textwidth,width=0.138\textwidth]{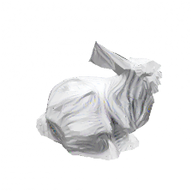}
        \includegraphics[height=0.138\textwidth,width=0.138\textwidth]{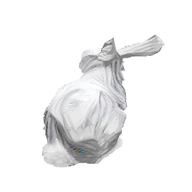}
        \includegraphics[height=0.138\textwidth,width=0.138\textwidth]{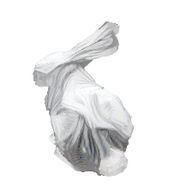} \\
    \end{center}
    \caption{Additional results of style transfer. The style images are {\it The Great Wave off Kanagawa}, (Hokusai, 1829-1832), {\it The Trial} (W. Lettl, 1981), {\it Bicentennial Print} (R. Lichtenstein, 1975), {\it Portrait of a Friend} (M. H. Maxy, 1926), {\it The Scream} (E. Munch, 1910), {\it Femme nue assise} (P. Picasso, 1909), and {\it Sketch}~\cite{johnson2016perceptual}.}
    \label{fig:appendix_style_transfer2}
\end{figure*}

\begin{figure*}[t]
    \begin{center}
        \includegraphics[height=0.138\textwidth,width=0.138\textwidth]{./figure/white.png}
        \includegraphics[height=0.138\textwidth,width=0.138\textwidth]{./images/teapot/rotation_00000000.png}
        \includegraphics[height=0.138\textwidth,width=0.138\textwidth]{./images/teapot/rotation_00000300.png}
        \includegraphics[height=0.138\textwidth,width=0.138\textwidth]{./images/teapot/rotation_00000240.png}
        \includegraphics[height=0.138\textwidth,width=0.138\textwidth]{./images/teapot/rotation_00000180.png}
        \includegraphics[height=0.138\textwidth,width=0.138\textwidth]{./images/teapot/rotation_00000120.png}
        \includegraphics[height=0.138\textwidth,width=0.138\textwidth]{./images/teapot/rotation_00000060.png} \\
        \includegraphics[height=0.138\textwidth,width=0.138\textwidth]{./images/style_transfer/bailly1.jpg}
        \includegraphics[height=0.138\textwidth,width=0.138\textwidth]{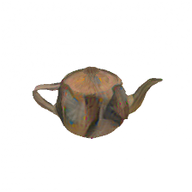}
        \includegraphics[height=0.138\textwidth,width=0.138\textwidth]{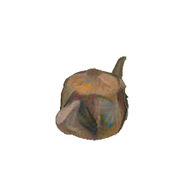}
        \includegraphics[height=0.138\textwidth,width=0.138\textwidth]{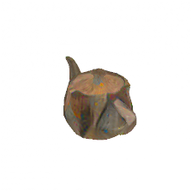}
        \includegraphics[height=0.138\textwidth,width=0.138\textwidth]{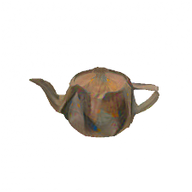}
        \includegraphics[height=0.138\textwidth,width=0.138\textwidth]{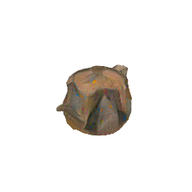}
        \includegraphics[height=0.138\textwidth,width=0.138\textwidth]{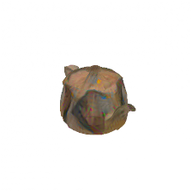} \\
        \includegraphics[height=0.138\textwidth,width=0.138\textwidth]{./images/style_transfer/coupland1.jpg}
        \includegraphics[height=0.138\textwidth,width=0.138\textwidth]{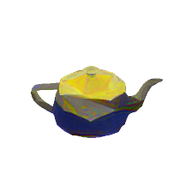}
        \includegraphics[height=0.138\textwidth,width=0.138\textwidth]{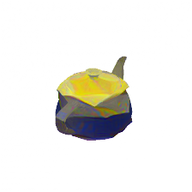}
        \includegraphics[height=0.138\textwidth,width=0.138\textwidth]{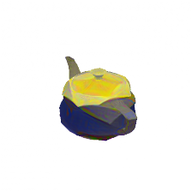}
        \includegraphics[height=0.138\textwidth,width=0.138\textwidth]{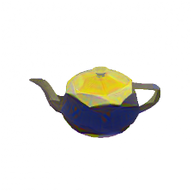}
        \includegraphics[height=0.138\textwidth,width=0.138\textwidth]{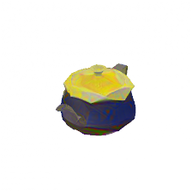}
        \includegraphics[height=0.138\textwidth,width=0.138\textwidth]{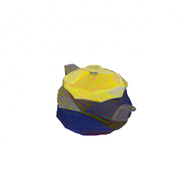} \\
        \includegraphics[height=0.138\textwidth,width=0.138\textwidth]{./images/style_transfer/elder1.jpg}
        \includegraphics[height=0.138\textwidth,width=0.138\textwidth]{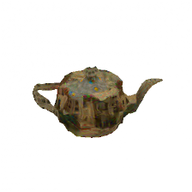}
        \includegraphics[height=0.138\textwidth,width=0.138\textwidth]{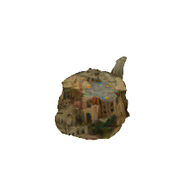}
        \includegraphics[height=0.138\textwidth,width=0.138\textwidth]{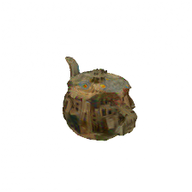}
        \includegraphics[height=0.138\textwidth,width=0.138\textwidth]{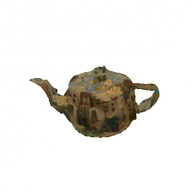}
        \includegraphics[height=0.138\textwidth,width=0.138\textwidth]{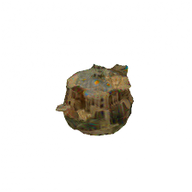}
        \includegraphics[height=0.138\textwidth,width=0.138\textwidth]{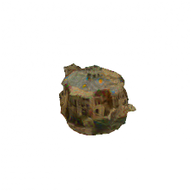} \\
        \includegraphics[height=0.138\textwidth,width=0.138\textwidth]{./images/style_transfer/feininger1.jpg}
        \includegraphics[height=0.138\textwidth,width=0.138\textwidth]{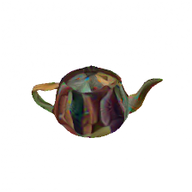}
        \includegraphics[height=0.138\textwidth,width=0.138\textwidth]{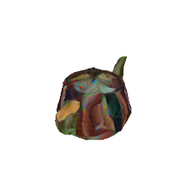}
        \includegraphics[height=0.138\textwidth,width=0.138\textwidth]{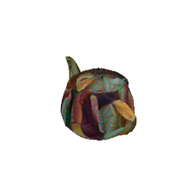}
        \includegraphics[height=0.138\textwidth,width=0.138\textwidth]{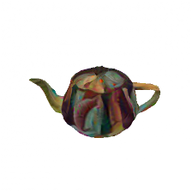}
        \includegraphics[height=0.138\textwidth,width=0.138\textwidth]{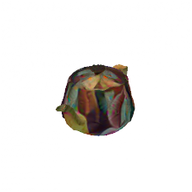}
        \includegraphics[height=0.138\textwidth,width=0.138\textwidth]{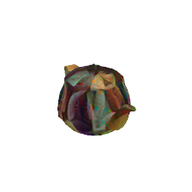} \\
        \includegraphics[height=0.138\textwidth,width=0.138\textwidth]{./images/style_transfer/gino1.jpg}
        \includegraphics[height=0.138\textwidth,width=0.138\textwidth]{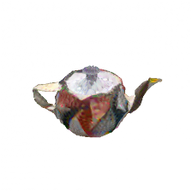}
        \includegraphics[height=0.138\textwidth,width=0.138\textwidth]{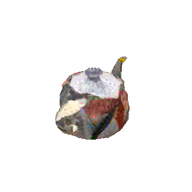}
        \includegraphics[height=0.138\textwidth,width=0.138\textwidth]{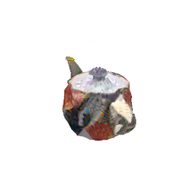}
        \includegraphics[height=0.138\textwidth,width=0.138\textwidth]{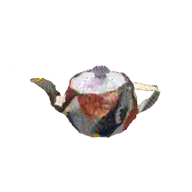}
        \includegraphics[height=0.138\textwidth,width=0.138\textwidth]{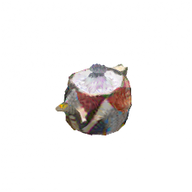}
        \includegraphics[height=0.138\textwidth,width=0.138\textwidth]{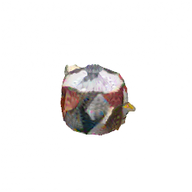} \\
        \includegraphics[height=0.138\textwidth,width=0.138\textwidth]{./images/style_transfer/gogh2.jpg}
        \includegraphics[height=0.138\textwidth,width=0.138\textwidth]{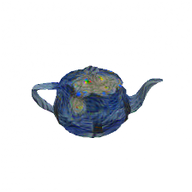}
        \includegraphics[height=0.138\textwidth,width=0.138\textwidth]{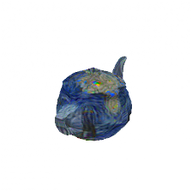}
        \includegraphics[height=0.138\textwidth,width=0.138\textwidth]{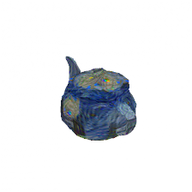}
        \includegraphics[height=0.138\textwidth,width=0.138\textwidth]{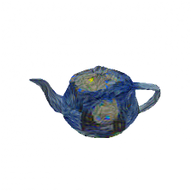}
        \includegraphics[height=0.138\textwidth,width=0.138\textwidth]{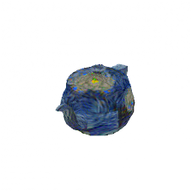}
        \includegraphics[height=0.138\textwidth,width=0.138\textwidth]{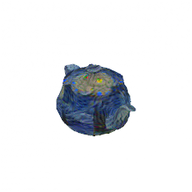} \\
        \includegraphics[height=0.138\textwidth,width=0.138\textwidth]{./images/style_transfer/gris1.jpg}
        \includegraphics[height=0.138\textwidth,width=0.138\textwidth]{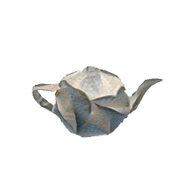}
        \includegraphics[height=0.138\textwidth,width=0.138\textwidth]{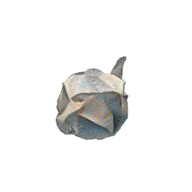}
        \includegraphics[height=0.138\textwidth,width=0.138\textwidth]{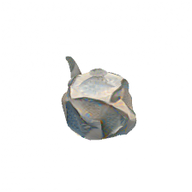}
        \includegraphics[height=0.138\textwidth,width=0.138\textwidth]{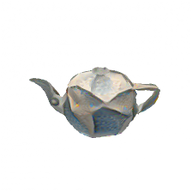}
        \includegraphics[height=0.138\textwidth,width=0.138\textwidth]{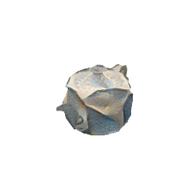}
        \includegraphics[height=0.138\textwidth,width=0.138\textwidth]{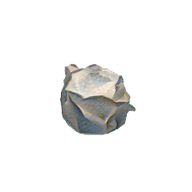} \\
    \end{center}
    \caption{Additional results of style transfer. The style images are {\it Self-Portrait} (A. Bailly, 1917), {\it Thomson No. 5 (Yellow Sunset)} (D. Coupland, 2011), {\it The Tower of Babel} (P. Bruegel the Elder, 1563), {\it Jesuits III} (L. Feininger, 1915), {\it Ritmo plastico del 14 luglio} (S. Gino, 1913), {\it The Starry Night} (V. van Gogh, 1889), and {\it Portrait of Pablo Picasso} (J. Gris, 1912).}
    \label{fig:appendix_style_transfer3}
\end{figure*}

\begin{figure*}[t]
    \begin{center}
        \includegraphics[height=0.138\textwidth,width=0.138\textwidth]{./figure/white.png}
        \includegraphics[height=0.138\textwidth,width=0.138\textwidth]{./images/teapot/rotation_00000000.png}
        \includegraphics[height=0.138\textwidth,width=0.138\textwidth]{./images/teapot/rotation_00000300.png}
        \includegraphics[height=0.138\textwidth,width=0.138\textwidth]{./images/teapot/rotation_00000240.png}
        \includegraphics[height=0.138\textwidth,width=0.138\textwidth]{./images/teapot/rotation_00000180.png}
        \includegraphics[height=0.138\textwidth,width=0.138\textwidth]{./images/teapot/rotation_00000120.png}
        \includegraphics[height=0.138\textwidth,width=0.138\textwidth]{./images/teapot/rotation_00000060.png} \\
        \includegraphics[height=0.138\textwidth,width=0.138\textwidth]{./images/style_transfer/hokusai1.jpg}
        \includegraphics[height=0.138\textwidth,width=0.138\textwidth]{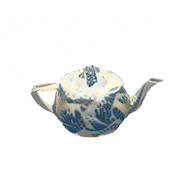}
        \includegraphics[height=0.138\textwidth,width=0.138\textwidth]{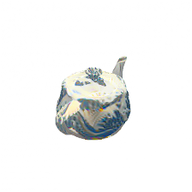}
        \includegraphics[height=0.138\textwidth,width=0.138\textwidth]{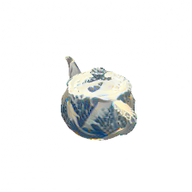}
        \includegraphics[height=0.138\textwidth,width=0.138\textwidth]{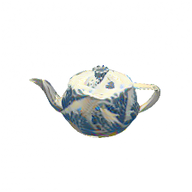}
        \includegraphics[height=0.138\textwidth,width=0.138\textwidth]{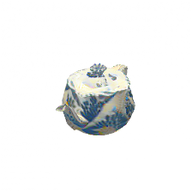}
        \includegraphics[height=0.138\textwidth,width=0.138\textwidth]{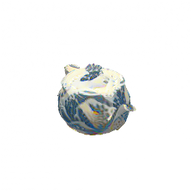} \\
        \includegraphics[height=0.138\textwidth,width=0.138\textwidth]{./images/style_transfer/lettl1.jpg}
        \includegraphics[height=0.138\textwidth,width=0.138\textwidth]{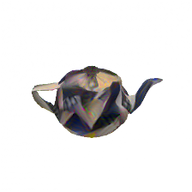}
        \includegraphics[height=0.138\textwidth,width=0.138\textwidth]{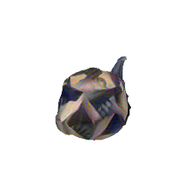}
        \includegraphics[height=0.138\textwidth,width=0.138\textwidth]{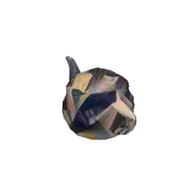}
        \includegraphics[height=0.138\textwidth,width=0.138\textwidth]{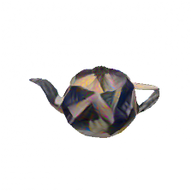}
        \includegraphics[height=0.138\textwidth,width=0.138\textwidth]{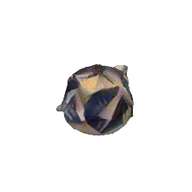}
        \includegraphics[height=0.138\textwidth,width=0.138\textwidth]{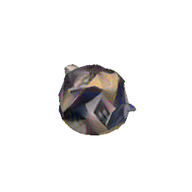} \\
        \includegraphics[height=0.138\textwidth,width=0.138\textwidth]{./images/style_transfer/lichtenstein1.jpg}
        \includegraphics[height=0.138\textwidth,width=0.138\textwidth]{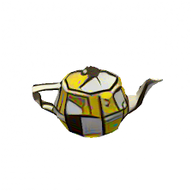}
        \includegraphics[height=0.138\textwidth,width=0.138\textwidth]{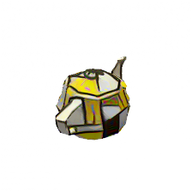}
        \includegraphics[height=0.138\textwidth,width=0.138\textwidth]{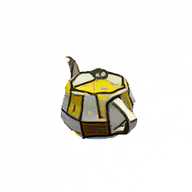}
        \includegraphics[height=0.138\textwidth,width=0.138\textwidth]{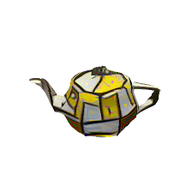}
        \includegraphics[height=0.138\textwidth,width=0.138\textwidth]{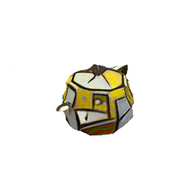}
        \includegraphics[height=0.138\textwidth,width=0.138\textwidth]{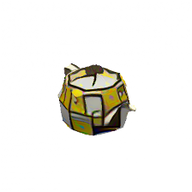} \\
        \includegraphics[height=0.138\textwidth,width=0.138\textwidth]{./images/style_transfer/maxy1.jpg}
        \includegraphics[height=0.138\textwidth,width=0.138\textwidth]{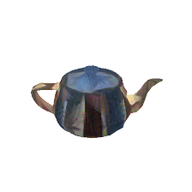}
        \includegraphics[height=0.138\textwidth,width=0.138\textwidth]{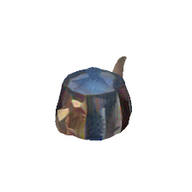}
        \includegraphics[height=0.138\textwidth,width=0.138\textwidth]{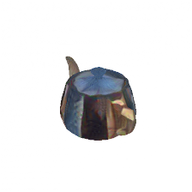}
        \includegraphics[height=0.138\textwidth,width=0.138\textwidth]{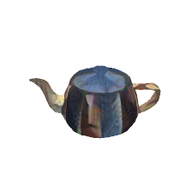}
        \includegraphics[height=0.138\textwidth,width=0.138\textwidth]{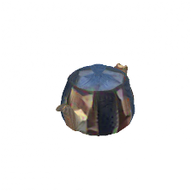}
        \includegraphics[height=0.138\textwidth,width=0.138\textwidth]{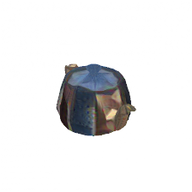} \\
        \includegraphics[height=0.138\textwidth,width=0.138\textwidth]{./images/style_transfer/munch1.jpg}
        \includegraphics[height=0.138\textwidth,width=0.138\textwidth]{./images/teapot_munch1_lc_5000000000_ltv_100000/rotation_00000000.png}
        \includegraphics[height=0.138\textwidth,width=0.138\textwidth]{./images/teapot_munch1_lc_5000000000_ltv_100000/rotation_00000300.png}
        \includegraphics[height=0.138\textwidth,width=0.138\textwidth]{./images/teapot_munch1_lc_5000000000_ltv_100000/rotation_00000240.png}
        \includegraphics[height=0.138\textwidth,width=0.138\textwidth]{./images/teapot_munch1_lc_5000000000_ltv_100000/rotation_00000180.png}
        \includegraphics[height=0.138\textwidth,width=0.138\textwidth]{./images/teapot_munch1_lc_5000000000_ltv_100000/rotation_00000120.png}
        \includegraphics[height=0.138\textwidth,width=0.138\textwidth]{./images/teapot_munch1_lc_5000000000_ltv_100000/rotation_00000060.png} \\
        \includegraphics[height=0.138\textwidth,width=0.138\textwidth]{./images/style_transfer/picasso1.jpg}
        \includegraphics[height=0.138\textwidth,width=0.138\textwidth]{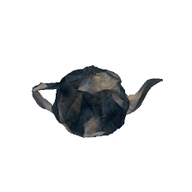}
        \includegraphics[height=0.138\textwidth,width=0.138\textwidth]{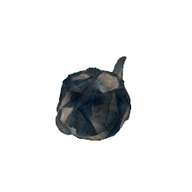}
        \includegraphics[height=0.138\textwidth,width=0.138\textwidth]{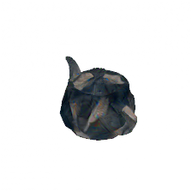}
        \includegraphics[height=0.138\textwidth,width=0.138\textwidth]{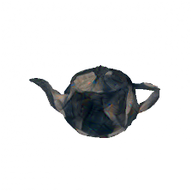}
        \includegraphics[height=0.138\textwidth,width=0.138\textwidth]{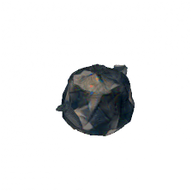}
        \includegraphics[height=0.138\textwidth,width=0.138\textwidth]{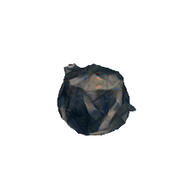} \\
        \includegraphics[height=0.138\textwidth,width=0.138\textwidth]{./images/style_transfer/sketch1.jpg}
        \includegraphics[height=0.138\textwidth,width=0.138\textwidth]{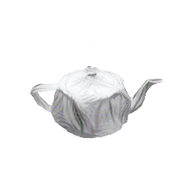}
        \includegraphics[height=0.138\textwidth,width=0.138\textwidth]{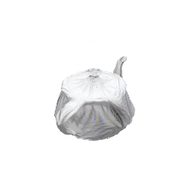}
        \includegraphics[height=0.138\textwidth,width=0.138\textwidth]{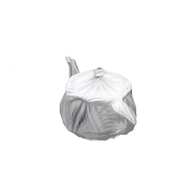}
        \includegraphics[height=0.138\textwidth,width=0.138\textwidth]{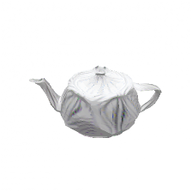}
        \includegraphics[height=0.138\textwidth,width=0.138\textwidth]{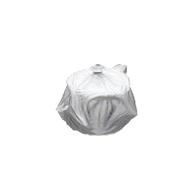}
        \includegraphics[height=0.138\textwidth,width=0.138\textwidth]{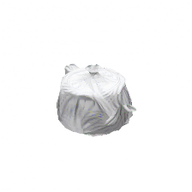} \\
    \end{center}
    \caption{Additional results of style transfer. The style images are {\it The Great Wave off Kanagawa}, (Hokusai, 1829-1832), {\it The Trial} (W. Lettl, 1981), {\it Bicentennial Print} (R. Lichtenstein, 1975), {\it Portrait of a Friend} (M. H. Maxy, 1926), {\it The Scream} (E. Munch, 1910), {\it Femme nue assise} (P. Picasso, 1909), and {\it Sketch}~\cite{johnson2016perceptual}.}
    \label{fig:appendix_style_transfer4}
\end{figure*}

Figure~\ref{fig:appendix_reconstruction1} and Figure~\ref{fig:appendix_reconstruction2} show additional results of 3D reconstruction. Figure~\ref{fig:appendix_style_transfer1}, Figure~\ref{fig:appendix_style_transfer2}, Figure~\ref{fig:appendix_style_transfer3}, and Figure~\ref{fig:appendix_style_transfer4} show additional results of style transfer.

% gs -dNOPAUSE -dBATCH -sDEVICE=pdfwrite -dCompatibilityLevel=1.4 -dPDFSETTINGS=/prepress -dColorImageResolution=1200 -dGrayImageResolution=1200 -dMonoImageResolution=1200 -sOutputFile=20171120.pdf paper.pdf

\end{document}